\documentclass[10pt,twocolumn,letterpaper]{article}

\usepackage{conf}
\usepackage{times}
\usepackage{epsfig}
\usepackage{graphicx}
\usepackage{amsmath}
\usepackage{amssymb}
\usepackage{subfig}
\newcommand*\diff{\mathop{}\!\mathrm{d}}
\usepackage[nolist]{acronym}

\usepackage{multirow}
\usepackage{booktabs}
\usepackage{colortbl}
\usepackage{xcolor}

\usepackage{fancyhdr}

\usepackage{arydshln}

\usepackage[pagebackref=true,breaklinks=true,letterpaper=true,colorlinks,bookmarks=false]{hyperref}

\conffinalcopy % *** Uncomment this line for the final submission

\ifconffinal\fi

\def\datasetname{iFF\xspace}
\def\numberofscenes{five\xspace}
\def\approach{NeRFtrinsic Four\xspace}

\definecolor{validgreen}{HTML}{97cc89}
\definecolor{invalidred}{HTML}{d98989}

\newcommand*\green{\cellcolor{validgreen}}
\newcommand*\red{\cellcolor{invalidred}}
\newcolumntype{P}[1]{>{\centering\arraybackslash}p{#1}}

\newcommand{\beginsupplement}{%
        \setcounter{table}{0}
        \renewcommand{\thetable}{S\arabic{table}}%
        \setcounter{figure}{0}
        \renewcommand{\thefigure}{S\arabic{figure}}%
        \setcounter{section}{0}
        \renewcommand{\thefigure}{S\arabic{section}}%
     }

\begin{document}
\begin{acronym}[Bspwwww.]  % Längstes Kürzel in der nachfolgenden
                       % Liste um die Breite der Spalte für die
                       % Abkürzungen zu bestimmen.

%% Eintrag: \acro{Referenzname}[Kürzel]{Langform}
%% Im Text wird die Abkürzung dann mit \ac{Referenzname} benutzt.
%Az
\acro{ar}[AR]{augmented reality}
\acro{ate}[ATE]{absolute trajectory error}
\acro{bvip}[BVIP]{blind or visually impaired people}
% C
\acro{cnn}[CNN]{convolutional neural network}
%F
\acro{fov}[FoV]{field of view}
%G
\acro{gan}[GAN]{generative adversarial network}
\acro{gcn}[GCN]{graph convolutional Network}
\acro{gnn}[GNN]{Graph Neural Network}
%H
\acro{hmi}[HMI]{Human-Machine-Interaction}
\acro{hmd}[HMD]{head-mounted display}
\acro{mr}[MR]{mixed reality}
% I
\acro{iot}[IoT]{internet of things}
% L
\acro{llff}[LLFF]{Local Light Field Fusion}
\acro{bleff}[BLEFF]{Blender Forward Facing}

\acro{lpips}[LPIPS]{learned perceptual image patch similarity}
%N
\acro{nerf}[NeRF]{neural radiance fields}
\acro{nvs}[NVS]{novel view synthesis}
% M
\acro{mlp}[MLP]{multilayer perceptron}
\acro{mrs}[MRS]{Mixed Region Sampling}

%O
\acro{or}[OR]{Operating Room}
%P
\acro{pbr}[PBR]{physically based rendering}
\acro{psnr}[PSNR]{peak signal-to-noise ratio}
\acro{pnp}[PnP]{Perspective-n-Point}
%Q
%R
%
\acro{sus}[SUS]{system usability scale}
\acro{ssim}[SSIM]{similarity index measure}
\acro{sfm}[SfM]{structure from motion}
\acro{slam}[SLAM]{simultaneous localization and mapping}

%T
\acro{tp}[TP]{True Positive}
\acro{tn}[TN]{True Negative}
\acro{thor}[thor]{The House Of inteRactions}
%U
\acro{ueq}[UEQ]{User Experience Questionnaire}
%V
\acro{vr}[VR]{virtual reality}
%W
\acro{who}[WHO]{World Health Organization}
\acro{ycb}[YCB]{Yale-CMU-Berkeley}
\acro{yolo}[YOLO]{you only look once}

\end{acronym}
%%%%%%%%% TITLE
\pagestyle{fancy}
\fancyhead{} % clear all header fields
\fancyfoot{} % clear all footer fields
\fancyfoot[LO]{\textbf{This is a preprint and the paper is under review at Computer Vision and Image Understanding.}}

\title{NeRFtrinsic Four: An End-To-End Trainable NeRF Jointly Optimizing Diverse Intrinsic and Extrinsic Camera Parameters}

\author{Hannah Schieber \\ %   %
\and Fabian Deuser \\ % %
\and Bernhard Egger \\
\and Norbert Oswald \\
\and Daniel Roth \\ %
\and \parbox{2.1in}{\centering Human-Centered Computing \\ and Extended Reality \\
Friedrich-Alexander Universität (FAU) \\ Erlangen-Nürnberg \\
Erlangen, Germany \\
{\tt\small hannah.schieber@fau.de, d.roth@fau.de}} \\
\and \parbox{2in}{\centering Lehrstuhl für Graphische Datenverarbeitung (LGDV) \\
Friedrich-Alexander Universität (FAU) \\ Erlangen-Nürnberg \\
Erlangen, Germany \\
{\tt\small bernhard.egger@fau.de}} \\
\and \parbox{1.8in}{\centering Institute for Distributed Intelligent Systems \\
University of the \\ Bundeswehr Munich \\
Munich, Germany \\
{\tt\small fabian.deuser@unibw.de, norbert.oswald@unibw.de}}
}

\maketitle% Remove page # from the first page of camera-ready.
\ifconffinal\thispagestyle{empty}\fi
% The preprint should include a statement that the paper is under consideration at Computer Vision and Image Understanding.

%%%%%%%%% ABSTRACT
\begin{abstract}
Novel view synthesis using neural radiance fields (NeRF) is the state-of-the-art technique for generating high-quality images from novel viewpoints. Existing methods require a priori knowledge about extrinsic and intrinsic camera parameters. This limits their applicability to synthetic scenes, or real-world scenarios with the necessity of a preprocessing step. Current research on the joint optimization of camera parameters and NeRF focuses on refining noisy extrinsic camera parameters and often relies on the preprocessing of intrinsic camera parameters. Further approaches are limited to cover only one single camera intrinsic. To address these limitations, we propose a novel end-to-end trainable approach called NeRFtrinsic Four. We utilize Gaussian Fourier features to estimate extrinsic camera parameters and dynamically predict varying intrinsic camera parameters through the supervision of the projection error. Our approach outperforms existing joint optimization methods on LLFF and BLEFF. In addition to these existing datasets, we introduce a new dataset called iFF with varying intrinsic camera parameters. NeRFtrinsic Four is a step forward in joint optimization NeRF-based view synthesis and enables more realistic and flexible rendering in real-world scenarios with varying camera parameters.
\end{abstract}

%%%%%%%%% BODY TEXT
\section{Introduction}

\begin{figure}[t!]
  \centering
  
\includegraphics[width=0.85\columnwidth]{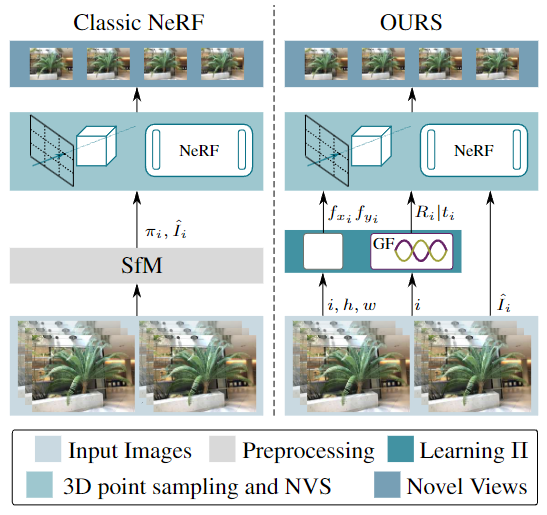}
  
  \caption{\textbf{The classic NeRF framework compared to our NeRFtrinsic Four.} Training a \acs{nerf} is usually limited to one type of camera and requires known camera parameters. We present \approach which jointly optimizes the camera parameters ($\Pi$) of multiple diverse cameras without the necessity of a preprocessing step. Our approach utilizes Gaussian Fourier features (GF) to learn the extrinsic camera parameters. Furthermore, we individually optimize the intrinsic camera parameters per given camera.
  }
\end{figure}

Generating novel views and producing rich, photo-realistic images requires multiple camera angles from different viewpoints to generate a detailed 3D scene representation. For the generation of novel views the knowledge of the intrinsic and extrinsic camera parameters of the training images is crucial~\cite{heigl_plenoptic_1999,levoy_light_1996,mildenhall_nerf_2020,szeliski_stereo_1998}. Intrinsic camera parameters that are dependent on camera properties like focal length or pixel dimensions impact the image that represents a captured part of a scene. In turn, individual camera parameters influence the projection of pixels into the 3D coordinate space. This projection into the 3D space is required by \ac{nerf}~\cite{mildenhall_nerf_2020, meng_gnerf_2021, wang_nerf_2021} which use the intrinsic parameters for the ray projection. Besides intrinsic camera parameters, the camera angle and position, denoted as extrinsic, are decisive.

Traditional approaches use rich \ac{sfm} algorithms like COLMAP~\cite{schonberger_structure--motion_2016} to determine  camera parameters in a preprocessing step and later utilize these estimations to train the \ac{nerf}~\cite{martin-brualla_nerf_2021,mildenhall_nerf_2020}. Although \ac{sfm} enables training, the preprocessing step is always necessary for new data. This prevents the \ac{nerf} from being end-to-end capable. Moreover, standard \ac{sfm} algorithms highly depend on texture to estimate accurate camera parameters. 

One use case is the representation of a 3D scene based on images from varying cameras. To easily generate a 3D representation of such a scenario, the network should be capable of processing the images directly without requiring any preprocessing. To avoid the preprocessing step, existing approaches investigate the joint optimization of camera parameters and the \ac{nerf}~\cite{lin_barf_2021,meng_gnerf_2021,yen-chen_inerf_2021,wang_nerf_2021,xia_sinerf_2022}. Concurrent joint optimization approaches either require given intrinsic camera parameters~\cite{lin_barf_2021,meng_gnerf_2021}, assume a pretrained \ac{nerf} as given~\cite{yen-chen_inerf_2021} or are restricted to one single camera~\cite{wang_nerf_2021,xia_sinerf_2022}.

We propose \approach, a novel approach that optimizes diverse intrinsic and extrinsic camera parameters along with \ac{nvs}. Unlike existing approaches, our method is not constrained to one single camera and does not require a preprocessing step to estimate the camera parameters. It also does not rely on a pretrained \ac{nerf}. We demonstrate the effectiveness of our approach on three benchmarks, namely \acs{llff}~\cite{mildenhall_local_2019}, \acs{bleff}~\cite{wang_nerf_2021} and our own \datasetname. Our evaluation shows that \approach outperforms state-of-the-art joint optimization methods in terms of image quality and camera parameter estimation on \acs{llff}, \acs{bleff} and \datasetname. Overall, our approach provides a more versatile solution for handling real-world scenes with varying cameras.

In summary, our approach contributes:

\begin{itemize}
    \item A dynamic joint end-to-end trainable optimization framework, capable of handling  diverse cameras.
    \item A pose-\ac{mlp}, using Gaussian Fourier features for the handling of challenging poses.   
    \item Our novel \datasetname dataset focusing on the challenge of diverse cameras, on which we demonstrate the advantages of \approach\footnote{\href{https://hannahhaensen.github.io/nerftrinsic_four/}{GitHub}}.
\end{itemize}
 
\section{Related Work}
Our approach addresses the joint optimization of \ac{nerf} as well as intrinsic and extrinsic camera parameters. We first review approaches, that predict camera parameters, followed by the combination of \ac{nerf} and camera pose refinement. Most importantly, we consider approaches estimating camera parameters and optimizing \ac{nerf} without prior initialization.

\subsection{Camera Parameter Estimation}
Traditional approaches which estimate the camera parameters often use \ac{sfm}. \ac{sfm} algorithms extract features and match them to find a 2D-3D correspondence. They estimate candidate poses and apply classical or optimized RANSAC to find the best matching poses~\cite{brachmann_visual_2021,fischler_random_1981}. 

In addition to \ac{sfm}-based approaches, \ac{slam} algorithms often jointly optimize the camera parameters together with the 3D reconstruction. MonoSLAM~\cite{davison_monoslam_2007} and ORB-SLAM~\cite{campos_orb-slam3_2021} estimate the  extrinsic camera parameters using feature correspondences. Others apply the photometric loss to optimize the camera pose~\cite{engel_lsd-slam_2014,forster_svo_2014,newcombe_dtam_2011}. COLMAP~\cite{schonberger_structure--motion_2016} is typically used to preprocess the camera parameters for \ac{nerf}. It first creates a sparse reconstruction using \ac{sfm}, and afterwards applies dense modelling using multi-view stereo.

Besides \ac{sfm} or \ac{slam}, deep learning based approaches are applied. Lee et al.~\cite{lee_camera--robot_2020} estimate the extrinsic camera parameters, assuming the intrinsic parameters to be known. The pose is estimated via the prediction of keypoints by a deep neural network using purely synthetic data. Elmoogy et al.~\cite{elmoogy_pose-gnn_2021} use a \ac{cnn} to extract features from each image and feed a \ac{gnn} to find the extrinsic camera parameters.  In addition to only predicting the extrinsic camera parameters, Butt et al.~\cite{butt_camera_2022} simultaneously estimate intrinsic camera parameters using a  \ac{cnn}. They apply Inception-v3 and a camera projection loss to estimate all camera parameters.

\subsection{\acs{nerf} and Camera Pose Refinement}
Mildenhall et al.~\cite{mildenhall_nerf_2020} introduced \ac{nerf}, which encodes a scene representation in a \ac{mlp}. 
Subsequent approaches optimize towards a higher image quality, handle fewer input views~(e.g.~\cite{barron_mip-nerf_2021,kulkarni_360fusionnerf_2022,martin-brualla_nerf_2021}) or use search engine results as input and handle the camera parameters via COLMAP~\cite{martin-brualla_nerf_2021}. 
Others investigate the impact of jointly optimizing extrinsic camera parameters and \ac{nerf} to improve \ac{nvs} quality~\cite{jeong_self-calibrating_2021,lin_barf_2021,meng_gnerf_2021,truong_sparf_2022}. 

Jeong et al.~\cite{jeong_self-calibrating_2021} tackle the joint extrinsic camera parameter and \ac{nvs} optimization with the geometric loss. This loss jointly optimizes the \ac{nerf} and extrinsic camera parameters focusing on forward-facing scenes. Besides optimizing initialized camera parameters, their approach can estimate unknown parameters by finding correspondences.
Lin et al.~\cite{lin_barf_2021} enhance this idea by using a coarse-to-fine annealing schedule in BARF. BARF jointly optimizes the 3D scene representation and registers the camera poses, initialized by noisy camera poses. Chen et al.~\cite{chng_gaussian_2022} improve BARF using Gaussian activation functions. While Truong et al.~\cite{truong_sparf_2022} also optimize on noisy poses, they additionally consider only sparse input views.

Apart from BARF-like approaches, Meng et al.~\cite{meng_gnerf_2021} combine \ac{nerf} and a GAN in GNeRF. GNeRF first randomly samples poses from a predefined pose sampling space and then optimizes the \ac{nerf}. The discriminator differentiates between real and fake images. An inversion network then learns the camera pose. Finally, the pose embedding and \ac{nerf} are optimized using the photometric loss.

\subsection{\acs{nerf} Without Known Camera Parameters}

While previous methods initialize their pose regression from noisy poses~\cite{lin_barf_2021,truong_sparf_2022,chng_gaussian_2022} or sample from a predefined space~\cite{meng_gnerf_2021}, iNerf~\cite{yen-chen_inerf_2021} optimizes the camera pose by inverting a pretrained \ac{nerf}. It starts from an initial pose and applies gradient descent to minimize the residual between the pixels in the rendered image and the observed image. Wang et al.~\cite{wang_nerf_2021} introduce \ac{nerf}$\text{-}\text{-}$ which optimizes not only the extrinsic but also the intrinsic camera parameters. The camera parameters are jointly optimized with the \ac{nerf} using the photometric loss. However, the intrinsic camera parameter estimation is limited to only one camera. They also limit their \ac{nerf} to focus on forward-facing scenes with a perturbation $\leq20^{\circ}$ and do not provide specific ray sampling for $360^{\circ}$ scenes. SiNeRF~\cite{xia_sinerf_2022} extends this approach by utilizing sinusoidal activation functions for radiance mapping and a novel \ac{mrs}. The \ac{mrs} ensures efficient training and prevents poor supervision from the lack of ray diversity. Their approach also focuses on one single camera intrinsic. Although they use \ac{mrs} the applicability is limited to forward-facing scenes. % \ac{nerf} minimizing the correspondence between pre-extracted poses

The most relevant work to ours is \ac{nerf}$\text{-}\text{-}$~\cite{wang_nerf_2021} and SiNeRF~\cite{xia_sinerf_2022} as they jointly optimize the \ac{nerf}, intrinsic camera parameters and extrinsic camera parameters, without any priors. However, the intrinsic camera parameters are restricted to one type of camera. Thus, we further optimize the camera intrinsic estimation to generalize on differing cameras at the same time. Moreover, we not only learn the six extrinsic parameters but show that a \ac{mlp} using Gaussian Fourier features can minimize the translation and rotation error and improves \ac{nvs} quality.

\section{Method}
\label{sec:preliminary}

In this work, we investigate the end-to-end trainable joint optimization of varying intrinsic  and extrinsic camera parameters as well as \acs{nvs}. 
The extrinsic camera parameters are learned by employing Gaussian Fourier feature mapping. For the intrinsic camera parameters, we apply dynamic parameter learning for each given camera. This allows \approach to learn independent intrinsic camera parameters for each camera. As depicted in Fig.~\ref{fig:overall_approach}, the output of extrinsic and intrinsic camera parameter estimation is then used for the \ac{nerf} training and jointly optimized with the \ac{nerf}. 

\subsection{Camera Parameters}
Following the definition of the pinhole camera model~\cite{hartley_multiple_2004}, $\Pi=K\left[R|t\right]$, the extrinsic camera parameters $\left[R|t\right]$ are used to convert from the world coordinate system to the camera coordinate system. The intrinsic camera parameters $K$ convert the points from the camera coordinate system into the image coordinate system. These parameters are the field of view, the focal length $f_x, f_y$, the principal point, skew and the geometric distortion. In summary, the camera parameters $\Pi$ influence which part of the scene around the camera is later part of the captured image. In turn, for \ac{nerf}, this defines the projection of rays and is essential for training  \ac{nerf}.

\subsubsection{Intrinsic Camera Parameters}
The number of possible images that are used for  \ac{nerf} is known a priori. For this, we construct a lightweight parameter array. The length of the array equals the number of varying cameras to optimize each image differentiated by their given camera towards their focal length. Thus, the images with varying height and width are independent from each other and independent $f = (f_x$, $f_y)$ values can be learned. Following the assumption of Wang et al.~\cite{wang_nerf_2021}, we consider the camera's principle points $c_x \approx W/2$ and $c_y \approx H/2$, where $H,~W$ denote the image height and width respectively. Thus, initialization is done individually for each intrinsic. The focal length parametrization, see equation~\ref{eq:focallength}, is a learned scaling factor $s_i$, which is initialized with $1.0$ for each camera $i$. Optimizing the square root of $s$, leads to better results in \ac{nerf}$\text{-}\text{-}$, which is why we adopted this 2nd-order trick. 
\begin{equation}
    \label{eq:focallength}
     \begin{split}
    f_{x_i} = s_{i}^{2}W,~
    f_{y_i} = s_{i}^{2}H
    \end{split}
\end{equation}

\begin{figure*}[t!]
  \centering
   \includegraphics[width=0.9\textwidth]{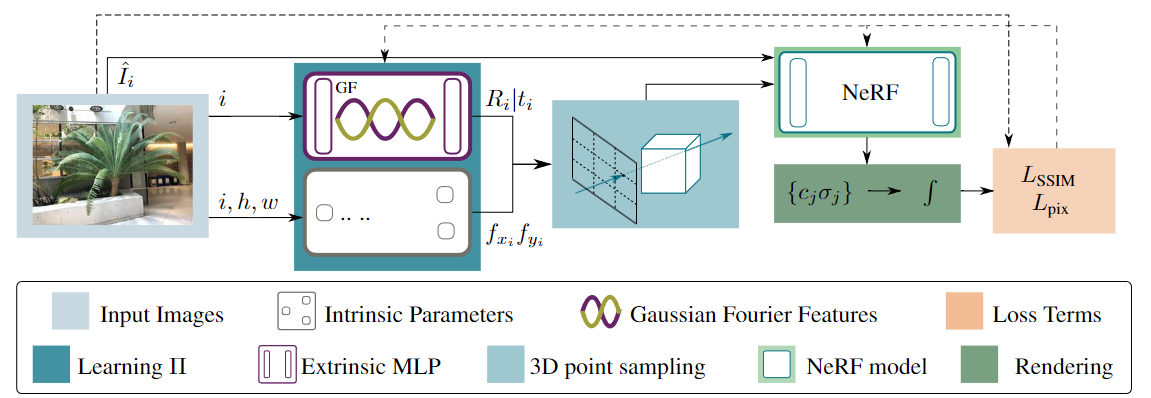}
  \caption{\textbf{Architecture of \approach, our end-to-end trainable approach.} We utilize Gaussian Fourier features (GF) in our pose \acs{mlp} to learn the extrinsic camera parameters and learn individual intrinsic camera parameters per given camera. We further stabilize the convergence of the pose \ac{mlp} with a \acs{ssim} loss function. Our framework is jointly optimized using the  \acs{ssim} loss ($L_{\text{SSIM}}$) and photometric loss ($L_{pix}$).\label{fig:overall_approach}}
\end{figure*}

\subsubsection{Gaussian Fourier Feature Mapping for Extrinsic Camera Parameters}

The extrinsic camera parameters, rotation $R \in SO(3)$ and translation $t \in \mathbb{R}^3$, are represented by a camera-to-world transformation matrix $\left[R|t\right]$ in $SE(3)$. We use  Gaussian Fourier feature mapping in a \ac{mlp} to learn these parameters. By utilizing different Fourier feature mappings, \acp{mlp} can learn high-frequency features from a low-dimensional input $\mathbf{v}$. Tancik et al.~\cite{tancik_fourier_2020} compared positional encoding given in equation~\ref{eq:posenc} and Gaussian Fourier feature mapping given in equation~\ref{eq:fourier}.

\begin{equation}\label{eq:posenc}
\gamma (\mathbf{v}) = \left[\text{cos}(2\pi\sigma^{j/m}\mathbf{v}), \text{sin}(2\pi\sigma^{j/m}\mathbf{v}) \right]^T\text{,} j=0,\text{...},m-1
\end{equation}

For both mappings, the scale $\sigma$ is determined through a hyperparameter search. While the positional encoding is deterministic, every value in matrix $\mathbf{B}$ for the Gaussian mapping is sampled from $\mathit{N}(0, \sigma^2)$.
Due to this sampling of random features, the bias towards axis aligned data, as in positional encoding, is avoided. Tancik et al. applied this only the image input to boost the performance of~\ac{nerf}.  %, but only applied to the image input

\begin{equation}\label{eq:fourier}
\gamma (\mathbf{v}) = \left [ \text{cos}(2\pi\mathbf{Bv}), \text{sin}(2\pi\mathbf{Bv}) \right ]^T, \mathbf{B} \in \mathbb{R}^{m\times d}
\end{equation}

We instead apply this for our extrinsic camera parameter estimation. The Gaussian Fourier feature mapping is used to map the index of each camera to a higher-dimensional space. As frequency parameter $m$ for the matrix $B$, we use 128 which results in an embedding size of 256. Our lightweight \ac{mlp} consists of three layers with a hidden size of $64$ and GELU activation functions. The embedding scale $\sigma$ is set to 10. 

\subsection{Neural Radiance Fields}

Using the camera parameters $\Pi$,  \ac{nerf} enable the generation of novel views. A scene in NeRF~\cite{mildenhall_nerf_2020} is represented as 5D vector function $f$ consisting of the 3D location $\textbf{x} = (x, y, z)$ and 2D viewing direction $\textbf{d} = (\theta, \phi)$ as input. A \ac{nerf} maps the 3D location $\textbf{x}$ and viewing direction $\textbf{d}$ to a radiance color $\textbf{c} = (r, g, b)$ and volume density $\sigma$, namely $f : \mathbb{R}^5\rightarrow\mathbb{R}^4$. The 5D coordinates are sampled along camera rays and the output of the \ac{mlp} is used in classical volumetric rendering techniques.

This differential volumetric rendering allows a fully optimizable pipeline for obtaining the pixel values based on the input coordinates. To render images from \ac{nerf}, the color from each pixel $\mathbf{p} = (u,v)$ on the image plane $\hat{I}_i$ is obtained by rendering the function $\mathcal{R}$, considering known camera parameters $\Pi$~\cite{mildenhall_nerf_2020,wang_nerf_2021}, see equation~\ref{eq:int}. 

\begin{equation}
\label{eq:int}
    \hat{I}_i (p) = \mathcal{R} (\mathbf{p},\pi_i|\Theta)) = \int_{h_n}^{h_f} T(\textit{h})\sigma(\mathbf{r}(h))\mathbf{c}(r(h),\mathbf{d})\diff h,
\end{equation}

The near and far bounds are denoted as $h_n$ (near) and $h_f$ (far)~\cite{wang_nerf_2021}. $\pi_i$ denotes the camera parameters and $T(h) = \text{exp}(-\int_{h_n}^{h_f}) \sigma(\mathbf{r}(s)ds)$ describes the accumulated transmission factor along the ray. To optimize the radiance field, \ac{nerf} minimize the mean squared error between rendered color and ground truth color also called photometric loss. In summary, the general \ac{nerf} framework can be formulated as $\Theta^{*} = \text{arg min} ~\mathcal{L} (\hat{I}|I, \Pi)$.

Wang et al.~\cite{wang_nerf_2021} adapt this framework to jointly optimize the \ac{nerf} as well as the intrinsic and extrinsic camera parameters, with the focus on forward-facing scenes. To achieve the joint optimization, Wang et al. reformulate the \ac{nerf} framework as denoted in equation~\ref{eq:nerfmm}. However, this framework is restricted to one single camera.

\begin{equation}
\label{eq:nerfmm}
    \Theta^{*}, \Pi^{*} = \text{arg min}~\mathcal{L} (I, PI | I)
\end{equation}

Our approach investigates the end-to-end trainable joint optimization of differing intrinsic and extrinsic camera parameters and \ac{nerf}. \approach learns varying intrinsic and extrinsic camera parameters along with the scene representation, see Fig.~\ref{fig:overall_approach}. As input, a set of RGB images $I$, potentially captured by varying cameras are used. We especially investigate intrinsic and extrinsic parameter optimization. Therefore, we adapt the \ac{nerf} framework, see equation~\ref{eq:approach}. Our framework considers differing intrinsic camera parameters $K^{*}_{\textit{cam}}$ and utilizes Gaussian Fourier features to predict the pose $\left[R|t\right]^{*}$ of each camera.

\begin{equation}
\label{eq:approach}
    \Theta^{*}, K^{*}_{\textit{cam}}, \left[R|t\right]^{*} = \text{arg min}~\mathcal{L} (\hat{I} \hat{K}_{\textit{cam}}, [\hat{R}| \hat{t}]| I)
\end{equation}

Consequently, we jointly optimize the intrinsic $K^{*}_{\textit{cam}}$ and extrinsic $\left[R|t\right]^{*}$ camera parameters along with the \ac{nerf}, while also allowing that images can be taken by different cameras ($\textit{cam}$). Our approach is depicted in Fig.~\ref{fig:overall_approach}, which shows the individual steps.

\section{Evaluation}

We evaluate the camera parameter estimation and \ac{nvs} quality of \approach on two real-world datasets, namely \ac{llff} and our intrinsic forward-facing (\datasetname) dataset. On \ac{llff}, we compare our approach in detail with the results from \ac{nerf}$\text{-}\text{-}$~\cite{wang_nerf_2021} and SiNeRF~\cite{xia_sinerf_2022}. To evaluate the performance of joint optimization approaches when diverse intrinsic camera parameters are present, we compare \ac{nerf}$\text{-}\text{-}$ and our approach on \datasetname. Additionally, we compare our results  with \ac{nerf}$\text{-}\text{-}$~\cite{wang_nerf_2021} on the synthetic \ac{bleff} dataset. 

\subsection{Datasets}

\textbf{LLFF:} Eight forward-facing scenes are included in \ac{llff}~\cite{mildenhall_local_2019}. These scenes have a varying number of images ranging from 20 up to 62. The pseudo ground truth of \ac{llff} derives from COLMAP.

\textbf{BLEFF:} Wang et al.~\cite{wang_nerf_2021} presented \acs{bleff} to evaluate camera parameter estimation accuracy and \acs{nvs} rendering quality. \acs{bleff} contains 14 scenes with 31 images each with a resolution of $1040 \times 1560$. For comparability, with \ac{nerf}$\text{-}\text{-}$ we followed the downscaling to a resolution of $520 \times 780$ and the $t010r010$ \acs{bleff} setup. 

\textbf{\datasetname:}  
We captured \numberofscenes real-world scenes namely T1, brick house, bike, fireplug and stormtrooper with 31 images each. The scenes were captured with an OAK-D Lite, an iPhone mini 13, and an iPad Air 2 with varying resolutions. Additionally, we added resizing, to show the influence of various image sizes. We received the pseudo ground truth by applying COLMAP. Details about the focal length and example images are included in the supplementary material. % Each scene of the dataset contains 31 images. 

\subsection{Evaluation Metrics}

To evaluate our approach, we consider two aspects. First, the rendering quality of the novel views. Here, we report \ac{psnr}, \ac{ssim}~\cite{wang_image_2004} and \ac{lpips}~\cite{zhang_unreasonable_2018}. The second aspect is the camera parameter estimation. To report on the intrinsic camera parameter quality, we measure the focal length error in pixels. For the extrinsic camera parameters, we use the  \ac{ate}~\cite{zhang_tutorial_2018,sturm_benchmark_2012} and similarity transformation $Sim(3)$ to align the ground truth and predicted poses. As metric, we report the rotation and translation error.

\subsection{Loss Function}

The commonly used loss $L_{pix} = \sum_{r \in R_i}^{}||I(r)- \hat{I}(r)||^2$, also called photometric loss, is applied to train the joined optimization.
% We conducted the photometric loss $L_{pix} = \sum_{r \in R_i}^{}||I(r)- \hat{I}(r)||^2$ which is commonly used to train a \ac{nerf}. 
Moreover, we found that the pose prediction stability can be strengthened by applying the \ac{ssim} loss $\mathit{L}_{\text{SSIM}}(P)=\frac{1}{N}\sum_{p \in P}^{} 1 - \text{SSIM}(p)$ during the starting phase of the training.

\begin{table}[t!]
    \centering
    \resizebox{\columnwidth}{!}{ 
    \begin{tabular}{l|cccc} \hline \hline
       Method  & \ac{psnr}$\uparrow$ & Focal Err. & Rot. Err. & Trans. Err  \\ 
        \hline\noalign{\smallskip}
       \textbf{\ac{llff}} (Real-World) \\ 
        ~~COLMAP~\cite{wang_nerf_2021} & 23.52 & - & - & - \\
        ~~OURS$^{COLMAP}$ & \textbf{23.70} & \textbf{10.4} & \textbf{0.87} & \textbf{0.002} \\ 
        \hdashline \noalign{\smallskip} 
        % \textbf{\ac{llff}} (Real-World), \\ 
       %\textbf{\ac{llff}} (Real-World) \\ 
        ~~\ac{nerf}$\text{-}\text{-}$~\cite{wang_nerf_2021} & 22.48 & 143.3 & 3.75 & 0.031 \\ 
        ~~SiNeRF~\cite{xia_sinerf_2022} & 21.49 & 155.2 & 14.81 & 0.033 \\
        ~~OURS & \textbf{23.14} & \textbf{103.7} & \textbf{2.10} & \textbf{0.008} \\ 
        \hline\noalign{\smallskip}
       \textbf{\ac{bleff}} (Synthetic) \\ 
        ~~COLMAP~\cite{wang_nerf_2021} &  \textbf{33.92} &  14.89 & 13.65 & \textbf{0.012} \\
        ~~\ac{nerf}$\text{-}\text{-}$~\cite{wang_nerf_2021} & 33.24&  20.55 & 4.45 & 0.065 \\
        ~~OURS & 33.57 &  \textbf{8.57} & \textbf{2.51} & 0.018 \\ 
        \hline\noalign{\smallskip} 
    \textbf{\datasetname} (Real-World) \\ 
    
        {~~COLMAP} & 24.95 & - & - & - \\
        ~~\ac{nerf}$\text{-}\text{-}$ & 23.78 & 206.27 & 8.91 &  0.159 \\
        {~~\ac{nerf}$\text{-}\text{-}$ + I} & 25.50 & 96.65 & 9.70 & \textbf{0.136} \\
        ~~OURS &  \textbf{27.15} & \textbf{75.14} & \textbf{5.46} & 0.191  \\ \hline \hline
    \end{tabular}
    }
    \caption{\textbf{Overview of the mean \ac{psnr} values and the camera parameter estimation errors on \ac{llff}, \ac{bleff} and \datasetname}. Our approach outperforms the existing joint optimization approaches on \ac{llff}, \ac{bleff} and \datasetname.} % When using COLMAP initialization and applying our method to finetune the camera parameters, we also outperform COLMAP-based \acs{nerf} on \acs{llff}.}
    \label{tab:overview}
\end{table}

\begin{table*}[t!]
   \begin{center}
       \resizebox{\textwidth}{!}{ 
    \begin{tabular}{l|ccc|ccc|ccc|ccc|ccc} \hline \hline
       \multirow{2}{*}{Scene} & \multicolumn{3}{c|}{PSNR$\uparrow$} & \multicolumn{3}{c|}{SSIM$\uparrow$} & \multicolumn{3}{c|}{LPIPS$\downarrow$} & \multicolumn{3}{c|}{Rot. Err.} & \multicolumn{3}{c}{Trans. Err.} \\  % \cmidrule(r){2.5-4.5}
        & \ac{nerf}$\text{-}\text{-}$ & SiNeRF & OURS & \ac{nerf}$\text{-}\text{-}$ &  SiNeRF & OURS & \ac{nerf}$\text{-}\text{-}$ & SiNeRF & OURS  & \ac{nerf}$--$& SiNeRF & OURS & \ac{nerf}$--$& SiNeRF & OURS \\ \hline 
       Fern     & 21.67 & 20.99 & \textbf{21.82} & 0.61 & 0.59 & \textbf{0.62} &  0.50 & 0.53 & \textbf{0.49} &  1.78 & \textbf{1.17} & 1.34 & 0.029  & \textbf{0.006} & \textbf{0.006} \\
       Flower    & 25.34 & \textbf{25.66} & 25.39 & 0.71 & \textbf{0.73} & 0.72 & \textbf{0.37} & \textbf{0.37} & \textbf{0.37} & 4.84 & 1.38 & \textbf{0.89} & 0.016 & \textbf{0.007} & \textbf{0.007} \\
       Fortress & 26.20 & 26.74 & \textbf{27.28} & 0.63 & 0.67 & \textbf{0.70} & 0.49 & 0.45 & \textbf{0.41} & 1.36 & 2.02 & \textbf{0.91} & 0.025 &  0.048 & \textbf{0.006} \\
       Horns     & 22.53 & 17.29 & \textbf{24.00} & 0.61 & 0.45 & \textbf{0.67} & 0.50 & 0.66 & \textbf{0.45} & 5.55 & 83.34 & \textbf{1.89} & 0.044 & 0.133 & \textbf{0.014} \\
       Leaves   & 18.88 & 17.38 & \textbf{18.97} & \textbf{0.53} & 0.43 & 0.50 & \textbf{0.47} & 0.52 & 0.49 & 3.90 & 14.46 & \textbf{2.65} & 0.016 & 0.100 & \textbf{0.005} \\
       Orchids   & 16.73 & 16.77 & \textbf{17.41} & 0.39 & 0.40 & \textbf{0.43} & 0.55 & 0.53 & \textbf{0.52} & 4.96 & 3.97 & \textbf{2.75} & 0.051 & 0.014 & \textbf{0.009} \\
       Room      & 25.84 & 24.84 & \textbf{27.12} & \textbf{0.84} & 0.80 & 0.82 & 0.44 & 0.51  &  \textbf{0.43}  & 2.77 & 4.92 & \textbf{1.33} & 0.030 & 0.022 & \textbf{0.006} \\
       T-Rex     & 22.67 & 22.14 & \textbf{22.92} & 0.72 & 0.68 & \textbf{0.74} & 0.44 & 0.49 & \textbf{0.43} & 4.67 & 7.19 & \textbf{3.90} & 0.036 & 0.027 & \textbf{0.008} \\ \hline
       Mean      & 22.48 & 21.49 & \textbf{23.14} & 0.63 & 0.60 &\textbf{0.66} & 0.47 & 0.51 & \textbf{0.44} & 3.73 & 14.81 & \textbf{2.10} & 0.031 & 0.033 & \textbf{0.008} \\ \hline \hline
    \end{tabular}
    }
   \end{center}
    % \caption{\textbf{Quantitative comparison between OURS, \ac{nerf}$\text{-}\text{-}$~\cite{wang_nerf_2021}, and SiNeRF~\cite{xia_sinerf_2022} on \ac{llff}}. We report \ac{psnr}, \ac{ssim} and \acs{lpips} to show the results on \ac{nvs}. For the extrinsic camera parameters we report the translation error and rotation error.}
    \caption{\textbf{Quantitative comparison between \approach (OURS), \ac{nerf}$\text{-}\text{-}$~\cite{wang_nerf_2021}, and SiNeRF~\cite{xia_sinerf_2022} on \ac{llff}}. For SiNeRF we retrained the approach with a layer dimension of 128, to ensure comparability with our approach and \ac{nerf}$\text{-}\text{-}$. We report \ac{psnr}, \ac{ssim} and \acs{lpips} to show the results on \ac{nvs}. For the extrinsic camera parameters we report the translation error and rotation error.}
    \label{tab:benchmark_llff}
\end{table*}

% For SiNeRF we retrained the approach with a layer dimension of 128, for comparability with our approach and \ac{nerf}$\text{-}\text{-}$.

\subsection{Implementation Details} 

\approach is implemented in PyTorch. It (a) excludes the hierarchical sampling strategy; (b) has a layer dimension of 128 instead of 256; (c) samples 1024 pixels
from each input image and 128 points along each ray~\cite{wang_nerf_2021}. To initialize our \ac{nerf}, we use Kaiming initialisation~\cite{he_delving_2015}. The focal length is initialized by the individual camera height and width considering the given resize factor of the individual dataset and optimized during training. The camera poses are initialized in $-z$ direction. 

We use Adam optimizer for the camera parameters and \ac{nerf}. The initial learning rate is set to $10^{-3}$ for all models. We decay the \ac{nerf} learning rate all 10 epochs by multiplying with 0.9954. The focal and pose learning rate are decayed every 100 epochs by multiplying 0.9. 

To ensure comparability with \approach, we retrain SiNeRF with a layer dimension of 128 as the original SiNeRF has a layer dimension of 256. For SiNeRF we tried 10 random seeds per scene and report the best results on \ac{llff}. In our supplementary material we show the comparison of \approach and SiNeRF with a layer dimension of 256.

\begin{figure}[t!]
    \centering
   % \subfloat{
    \includegraphics[width=\columnwidth]{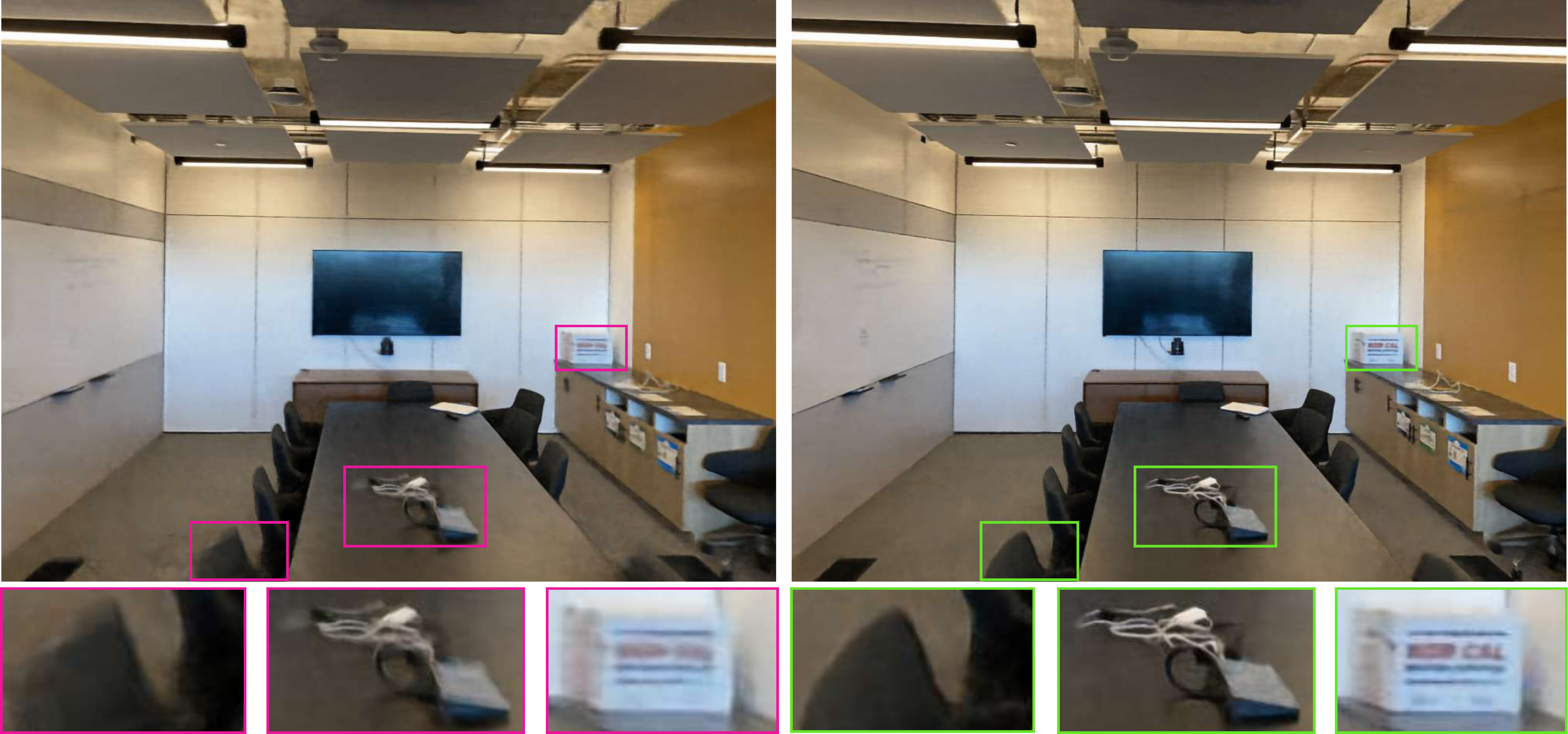}
       
   % \subfloat{
   %  \includegraphics[width=\columnwidth]{figures/fortress.pdf}
   % }
    \caption{\textbf{Visual results on \acs{llff}.} As shown in the scene room, our approach (right) gets finer details compared to \ac{nerf}$\text{-}\text{-}$ (left). This can be seen, for example, when looking at the chair in the front, the cables on the table or the box in the back on right side of the room.}
    \label{fig:fortress_fig}
\end{figure}

\subsection{Novel View Synthesis Quality}
We compare our approach with COLMAP-based \ac{nerf}, on \ac{llff} and \ac{bleff}. The COLMAP-based \acs{nerf} follows the same 128-layer dimension as \ac{nerf}$\text{-}\text{-}$. This results in lower \ac{nvs} quality compared to the vanilla \ac{nerf}~\cite{mildenhall_nerf_2020}.

As shown in Table~\ref{tab:overview}, for \acs{llff} and \acs{bleff} we achieve competitive \ac{nvs} compared to COLMAP without any need for preprocessing and outperform existing joint optimization approaches. On \ac{llff}, we outperform these approaches in \ac{psnr}, \ac{ssim} and \ac{lpips}. When using COLMAP initialization for the joint optimization we also outperform COLMAP-based \ac{nerf}. Detailed results for the COLMAP initialization can be found in the supplementary material. The detailed results on \ac{llff} for \acs{nvs} quality can be found in Table~\ref{tab:benchmark_llff}. We outperform \ac{nerf}$\text{-}\text{-}$ on all scenes on \ac{llff} and SiNeRF on seven out of eight scenes in \ac{psnr}. Overall we achieve better results for mean \ac{psnr}, mean \acs{ssim} and mean \acs{lpips} compared to all other joint optimization methods. 
We also report visual results in Fig.~\ref{fig:fortress_fig}, comparing our result on \ac{nvs} to \ac{nerf}$\text{-}\text{-}$. As shown in the room scene of \ac{llff}, we better reconstruct fine-grained details in the focus of the image and in the corners. 

Moreover, we outperform \ac{nerf}$\text{-}\text{-}$ on \ac{bleff} in \ac{psnr}, and perform equally in \ac{ssim}. The results for each scene are reported in Table~\ref{tab:benchmark_bleff}. 

On \datasetname we compared \ac{nerf}$\text{-}\text{-}$ and \approach (OURS), as shown in Table~\ref{tab:benchmark_iff}. Here, we first tested only our intrinsic camera prediction module, as \datasetname contains diverse focal lengths. Using our intrinsic camera module alone, we achieved an improved \ac{nvs} quality. Thereby, we already outperformed \ac{nerf}$\text{-}\text{-}$ in \ac{psnr} and \ac{ssim}. Combined with our Gaussian Fourier feature mapping for the extrinsic camera prediction we achieved an even better \ac{nvs} quality. As shown in Table~\ref{tab:benchmark_iff}, we outperform \ac{nerf}$\text{-}\text{-}$ on all scenes in \ac{psnr} and \ac{ssim}.

\begin{table*}
    \centering
       \resizebox{\textwidth}{!}{ 
    \begin{tabular}{|c|p{25pt}p{25pt}|p{25pt}p{25pt}|p{25pt}p{25pt}|p{25pt}p{25pt}|p{25pt}p{25pt}|p{25pt}p{25pt}|p{25pt}p{25pt}|p{25pt}p{25pt}|} \hline
    Skip/Scene & \multicolumn{2}{c|}{Fern} &  \multicolumn{2}{c|}{Flower}  &  \multicolumn{2}{c|}{Fortress}  & \multicolumn{2}{c|}{Horns} & \multicolumn{2}{c|}{Leaves}  &  \multicolumn{2}{c|}{Orchids}  &  \multicolumn{2}{c|}{Room}  &  \multicolumn{2}{c|}{T-Rex} \\ \hline
     & \scriptsize{\ac{nerf}$\text{-}\text{-}$}  & \scriptsize{OURS}  & \scriptsize{\ac{nerf}$\text{-}\text{-}$}  & \scriptsize{OURS} & \scriptsize{\ac{nerf}$\text{-}\text{-}$}  & \scriptsize{OURS} & \scriptsize{\ac{nerf}$\text{-}\text{-}$}  & \scriptsize{OURS} & \scriptsize{\ac{nerf}$\text{-}\text{-}$}  & \scriptsize{OURS} & \scriptsize{\ac{nerf}$\text{-}\text{-}$}  & \scriptsize{OURS} & \scriptsize{\ac{nerf}$\text{-}\text{-}$}  & \scriptsize{OURS} & \scriptsize{\ac{nerf}$\text{-}\text{-}$}  & \scriptsize{OURS} \\
    2 & \green & \green  & \green & \green & \green & \green & \green & \green & \green & \green & \green & \green & \green & \green  & \green & \green\\
    3 & \green & \green  & \green & \green & \red & \green & \green & \green & \green & \green & \red & \green & \red & \red  & \green & \red\\
    4 & \red & \green  & \green & \green & \red & \green & \red & \green & \green & \green & \green & \green & \red & \green  & \green & \green\\
    5 & \red & \red  & \green & \green & \red & \red & \red & \green & \green & \red & \red & \green & \red & \red  & \green & \green\\ \hline
    % 6 & \red & \red  & \green & \green & \red & \red & \green & \green & \red & \green & \red & \red & \green & \green  & \green & \green\\ \hline

    \end{tabular}
    }
    \caption{\textbf{Breaking point analysis of our Gaussian Fourier feature-based pose \acs{mlp} vs. \ac{nerf}$\text{-}\text{-}$ on \acs{llff}.} We used  every second, third, fourth, fifth or sixth image during training. A rotation error below  $20^{\circ}$ is considered as success (green). We outperform \ac{nerf}$\text{-}\text{-}$ as we succeed on 26 scenes and \ac{nerf}$\text{-}\text{-}$ only on 20.\label{tab:breaking_ana}}
\end{table*}

\begin{table*}[t!]
    \centering

   \begin{center}
    \resizebox{0.85\textwidth}{!}{ 
    \begin{tabular}{l|cc|cc|cc|cc|cc} \hline \hline
       \multirow{2}{*}{Scene} & \multicolumn{2}{c|}{\acs{psnr}} & \multicolumn{2}{c|}{\acs{ssim}} & \multicolumn{2}{c|}{Focal. Err.} & \multicolumn{2}{c|}{Rot. Err.} & \multicolumn{2}{c}{Trans. Err.} \\
       &  \ac{nerf}$\text{-}\text{-}$ & ~~OURS~~ & \ac{nerf}$\text{-}\text{-}$ & ~~OURS~~ &  \ac{nerf}$\text{-}\text{-}$ & ~~OURS~~ & \ac{nerf}$\text{-}\text{-}$ & ~~OURS~~   & \ac{nerf}$\text{-}\text{-}$ & ~~OURS~~\\ \hline 
        Airplane~~ & \textbf{30.57} & 29.99 & 0.83 &  \textbf{0.86} & 0.87 & 1.77 &  \textbf{0.61} & 1.04 &  \textbf{0.003} & 0.015 \\
        Balls  & 32.12 &  \textbf{34.39} & 0.81 & 0.64 & 15.44 & \textbf{14.53} & 13.43 & \textbf{0.81} & 0.285 & 0.004 \\
        Bathroom~~ & \textbf{31.58} & 31.28 & 0.94 & 0.92 & 0.39 & \textbf{0.19} & 1.50 & 1.49  & 0.004 & 0.001 \\
        Bed  & 32.41 &  \textbf{33.82} & 0.94 & \textbf{0.95} & 0.39 & \textbf{0.02} & \textbf{2.21} &  \textbf{2.21} & 0.004 &  \textbf{0.001} \\
        Castle &  \textbf{32.74} & 31.86 &  \textbf{0.89} & 0.87 & \textbf{3.23} & 16.06 & \textbf{3.17} & 4.41 &  \textbf{0.020} & 0.053 \\
        Chair  &  \textbf{32.24} & 31.54 &  \textbf{0.81} & \textbf{0.81} & 6.12 & \textbf{5.74} &  \textbf{3.52} & 5.44 & 0.078 &  \textbf{0.006}  \\
        Classroom~~ &  \textbf{25.14} & 24.90 & 0.86 &  \textbf{0.88} & 2.29 & \textbf{0.12} & 8.14 &  \textbf{5.38} & 0.032 &  \textbf{0.004} \\
        Deer &  \textbf{42.01} & 41.63 &  \textbf{0.99} &  \textbf{0.99} & \textbf{6.29} & 31.46 & 6.17 &  \textbf{5.51} & 0.166 &  \textbf{0.001}  \\
        Halloween & 29.30 &  \textbf{32.11} & 0.91 &  \textbf{0.94} & 9.77 & \textbf{6.53} & 6.74 &  \textbf{0.97} &  \textbf{0.022} & 0.030 \\
        Jugs &  \textbf{42.5} & 42.09 &  \textbf{0.99} &  \textbf{0.99} & \textbf{0.16} & 0.99 & 2.30 &  \textbf{0.72} & 0.065 &  \textbf{0.033} \\
        Root & 35.45 &  \textbf{37.00} & 0.97 &  \textbf{0.98} & 36.42 & \textbf{19.54} & 4.51 &  \textbf{0.45} & 0.100 &  \textbf{0.016} \\
        Roundtable & 39.88 &  \textbf{39.91} &  \textbf{0.99} & 0.98 & 206.10 & \textbf{17.92} & 9.68 &  \textbf{2.72} & 0.139 &  \textbf{0.020} \\
        Stone &  \textbf{31.74} & 31.36 &  \textbf{0.86} & 0.85 & \textbf{0.11} & 3.12 & \textbf{0.12} & 0.14 &  \textbf{0.001} &  \textbf{0.001} \\
        Valley & 27.66 &  \textbf{27.92} &  \textbf{0.75} & 0.74  & \textbf{0.05} & 1.82 & 0.25 &  \textbf{0.15} &  \textbf{0.001} &  \textbf{0.001}\\ \hline
        Mean & 33.24 &  \textbf{33.57} &  \textbf{0.90} &  \textbf{0.90} & 20.55 & \textbf{8.57} & 4.45 &  \textbf{2.15} & 0.064 &  \textbf{0.018} \\ \hline \hline
    \end{tabular}
    }
   \end{center}
    \caption{\textbf{Quantitative comparison of \ac{nerf}$\text{-}\text{-}$~\cite{wang_nerf_2021} and \approach (OURS) on \acs{bleff}.} We report \ac{psnr}, \ac{ssim}, translation error and rotation error. We outperform  \ac{nerf}$\text{-}\text{-}$ in \ac{psnr}, focal length error, rotation error and translation error.}
    \label{tab:benchmark_bleff}
\end{table*}
\subsection{Camera Parameter Estimation}

\subsubsection{Extrinsic Camera Parameters}
For each scene in \ac{llff}, we report the rotation and translation error of \ac{nerf}$\text{-}\text{-}$, SiNeRF$128$ and our approach in Table~\ref{tab:benchmark_llff}. 
We outperform both approaches with \approach. Furthermore, we show the advantage of the \acs{ssim} loss, exemplarily for the scenes in \ac{llff} in our supplementary material. For \ac{bleff} and \datasetname, we report the detailed results of \ac{nerf}$\text{-}\text{-}$ and \approach in Table~\ref{tab:benchmark_bleff} and in Table~\ref{tab:benchmark_iff}. Our approach shows an improved extrinsic camera parameter estimation on all datasets.

\paragraph{Positional Encoding}
Approaches like BARF~\cite{lin_barf_2021}  recognize the benefits of Fourier features for noisy pose estimation. However, they do not use Gaussian Fourier features but instead the so-called positional encoding. Our experiments show that the use of Gaussian Fourier features in comparison leads to a better convergence. For this purpose, we train our network with Gaussian Fourier features and positional encoding with ten different random seeds on each scene. The Gaussian Fourier features lead to an average rotation error of \textit{26.91} compared to the positional encoding with an average rotation error of \textit{45.95}. While the gap between the translation error remains small, the Gaussian Fourier features still outperform the positional encoding with an error of 0.039 compared to 0.042. The high error is caused by mirror poses in the ten runs. The runs are included in the score calculation to show that mirror poses are less likely with Gaussian Fourier features. A detailed overview of the error in each scene can be found in the supplementary material. 

\paragraph{Breaking Point Analysis}
A breaking point analysis for our pose \acs{mlp} shows an improved stability on the pose predictions by using fewer images. We compare our approach with \ac{nerf}$\text{-}\text{-}$ on \acs{llff}. A rotation error greater than $20^{\circ}$ is considered as failed. For the training, we used every second, third, fourth, fifth and sixth image. As shown in Table~\ref{tab:breaking_ana}, we succeed in 26 scenes, while \ac{nerf}$\text{-}\text{-}$ only succeeds in 20. % Additionally, breaking-point analysis on \datasetname can be found in the supplementary material. There we succee

\begin{table*}[t!]
    \centering

   \begin{center}
    \resizebox{\textwidth}{!}{ 
    \begin{tabular}{l|cccc|cccc|ccc|ccc|ccc} \hline \hline
    % & &&&&&& \\
       \multirow{2}{*}{Scene} & \multicolumn{4}{c|}{\acs{psnr}} & \multicolumn{4}{c|}{\acs{ssim}} & \multicolumn{3}{c|}{Focal Err.} & \multicolumn{3}{c|}{Rot Err.} & \multicolumn{3}{c}{Trans Err.}\\
        & {\ac{nerf}} & \ac{nerf}$\text{-}\text{-}$ & \ac{nerf}$\text{-}\text{-}$ + I & I+GF & {\ac{nerf}} & \ac{nerf}$\text{-}\text{-}$ & \ac{nerf}$\text{-}\text{-}$ + I & I+GF  & \ac{nerf}$\text{-}\text{-}$ & I & I+GF & \ac{nerf}$\text{-}\text{-}$ & I & I+GF & \ac{nerf}$\text{-}\text{-}$ & I & I+GF\\ \hline  
        T1 & {25.91} & 24.36 & 26.22 & \textbf{26.77} & {0.83} & 0.82 & 0.84 & \textbf{0.85} & 214.36 & \textbf{55.89} & 70.24 & 5.97 & 9.45 & 2.30 & 0.120 & \textbf{0.018} & 0.042\\ 
        Brick House & {\textbf{25.44}} & 21.09 & 23.98 & 24.91 & {\textbf{0.74}} & 0.66 & 0.71 & \textbf{0.71} & 254.73 & \textbf{67.80} & 89.99 & \textbf{4.53} & 4.49 & 5.44 & 0.075 & 0.090 &  \textbf{0.052} \\
        Fireplug & {22.60} & 22.52 & 22.95 & \textbf{25.15} & {0.65} & 0.61 & 0.63 & \textbf{0.71} & 267.89 & 123.24 & \textbf{92.14} & \textbf{5.45} & 5.61 & 5.51 & 0.006 & \textbf{0.001} & 0.007 \\
        Bike & {25.81} & 15.38 & 17.72 & \textbf{22.12} & {0.84} & 0.32 & 0.55 & \textbf{0.70} & 173.21 & 132.38 & \textbf{68.93} & 20.77 & 19.99 &  \textbf{6.58} & 0.042 & \textbf{0.035} & 0.060 \\
        Stormtropper & {25.00} & 36.23 & 36.61 & \textbf{36.67} & {0.90} & \textbf{0.97} & \textbf{0.97} & 0.99 & 121.18 & 103.93 & \textbf{51.51}  & 7.83 & 8.97 & \textbf{7.47} & 0.550 & \textbf{0.540} & 0.796\\ \hline
        Mean   & {24.95} &  23.92 & 25.50 & \textbf{27.12} & {0.79} & 0.68 & 0.74 & \textbf{0.80} & 206.27 & 96.65 & \textbf{74.56}  & 8.91 & 9.70 & \textbf{5.46} & 0.159 & \textbf{0.136} & 0.191\\ \hline \hline
    \end{tabular}
   }
   \end{center}
    \caption{\textbf{Quantitative comparison between {COLMAP-based} \ac{nerf}, \ac{nerf}$\text{-}\text{-}$~\cite{wang_nerf_2021}, \ac{nerf}$\text{-}\text{-}$ and our improved intrinsic estimation (I) and \approach (I+GF) on \datasetname.} We report \acs{psnr}, \acs{ssim}, rotation error, translation error and focal length error.}
    \label{tab:benchmark_iff}
\end{table*}

\subsubsection{Intrinsic Camera Parameters}

\begin{figure*}[t!]
    \centering
    \includegraphics[width=\textwidth]{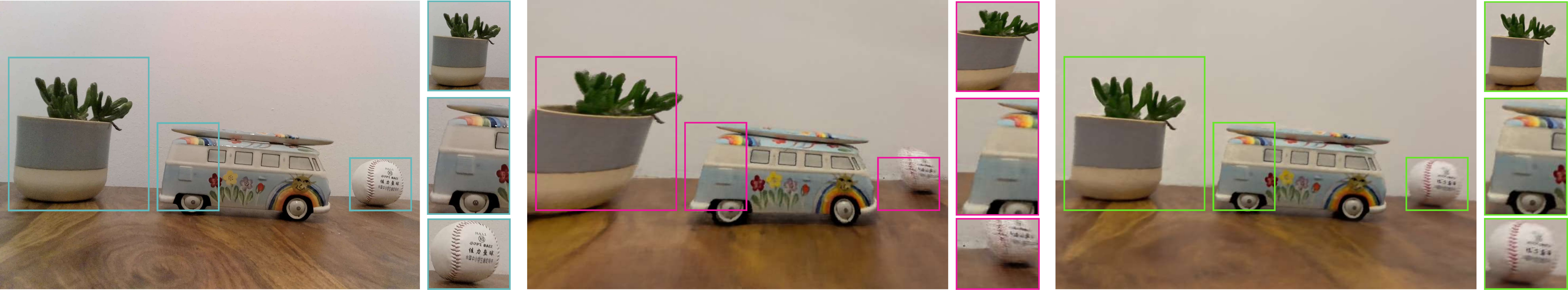}
    
    \caption{\textbf{Visual results on \datasetname.} As shown in scene T1 (left), our approach (right) provides a more accurate image compared to \ac{nerf}$\text{-}\text{-}$ (center). For example, the rear part of the car and the plant show a more accurate result with \approach. This demonstrates the influence of a correct focal length prediction when comparing both approaches.}
    \label{fig:bulli_fig}
\end{figure*}

In the focal length estimation, we outperform existing approaches on \ac{llff} and \ac{bleff}, see Table~\ref{tab:overview}. This indicates, that the joint optimization and an improved extrinsic camera parameter estimation also supports the intrinsic camera parameter estimation. As described in detail in the supplementary material, we outperform \ac{nerf}$\text{-}\text{-}$ on five out of eight scenes on \ac{llff} in the focal length estimation. On \ac{bleff} we achieve a better focal length estimation on eight out of 14 scenes, see Table~\ref{tab:benchmark_bleff}.  The scenes in both datasets are captured by a single camera. For this reason, we introduce \datasetname. The focal lengths in the scenes vary as we used different cameras and rescaling factors. While \ac{nerf}$\text{-}\text{-}$ averages the scaling factor over all images, we learn the individual intrinsic camera parameter per given camera. To show the influence of varying camera parameters, we trained \ac{nerf}$\text{-}\text{-}$, \ac{nerf}$\text{-}\text{-}$ combined with our intrinsic camera parameter method and our \approach on \datasetname. As presented in Table~\ref{tab:benchmark_iff}, the network better estimates the focal lengths. Using only our intrinsic module we already show an improved focal length estimation.  \approach predicts the focal length even better, possibly due to the improved prediction of the camera pose. A more accurate focal length has an impact on the results, as demonstrated in Fig.~\ref{fig:bulli_fig}. Unlike the image produced by \ac{nerf}$\text{-}\text{-}$, the image produced by \approach shows no distortion and is more accurate compared to the ground truth.

\section{Discussion}
\label{sec:discussion}

Our breaking point analysis shows that we outperform \ac{nerf}$\text{-}\text{-}$ on \ac{llff} and \datasetname. However, on some scenes our method needs a sufficient number of images for successful joint optimization, as indicated in Table~\ref{tab:breaking_ana} and the supplementary material. When dealing with multiple intrinsic camera parameters on \datasetname, the same conclusion emerges. Nevertheless, our approach can handle diverse intrinsic camera parameters while \ac{nerf}$\text{-}\text{-}$ learns only an average over the input. Fig.~\ref{fig:bulli_fig} shows that accurately estimated intrinsic camera parameters are crucial for a valid image representation. While an average over the images can lead to high distortions, our precise estimation of differing intrinsic camera parameters supports accurate image synthesis.

We also observe that a rotation error of $180^{\circ}$ leads to a mirror pose. This was already reported by Wang et al.~\cite{wang_nerf_2021}. We provide an example image in the supplementary material. While mirror poses still lead to a high \ac{psnr} value, the position of objects in the scene is shifted. To further stabilize the pose \ac{mlp}, we added the \ac{ssim} loss. Our supplementary material shows that our approach with the \ac{ssim} loss achieves better convergence for a fixed random seed. However, using the loss over the whole training process leads to a degradation in performance for the \ac{psnr} value.

Our experiments highlight the importance of accurate intrinsic and extrinsic camera parameter estimations for an end-to-end trainable \ac{nerf}. Incorrect extrinsic camera parameters can shift objects in the scene, while incorrect intrinsic camera parameters can distort the resulting image.
\section{Limitations}

A limitation of our work, inherited from previous research~\cite{wang_nerf_2021,xia_sinerf_2022}, is the focus on forward-facing scenes. This means that our approach is not suitable for $360^{\circ}$ scenes. While current approaches that can handle non-forward-facing scenes exist~\cite{lin_barf_2021,truong_sparf_2022,yen-chen_inerf_2021}, they typically rely on noisy poses or a pretrained \ac{nerf} for initialization, which are not transferable to real-world scenarios. Additionally, we found that predicting extrinsic camera parameters becomes more challenging in our joint optimization approach when dealing with varying camera intrinsic parameters.

Moreover, the initialization of the pose \ac{mlp} is challenging. Therefore, we see potential in a regularization method for  coordinate based \acp{mlp}~\cite{ramasinghe_regularizing_2022}.

\section{Conclusion}

We introduce \approach, an end-to-end trainable \ac{nerf} that jointly optimizes the extrinsic camera parameters, the intrinsic camera parameters, and the scene representation. Unlike other joint optimization frameworks, it is not limited to one type of camera. 
To validate our approach, we present our real-world \datasetname dataset, which demonstrates that \approach achieves better results compared to \ac{nerf}$\text{-}\text{-}$ when a diverse set of cameras is used. Additionally, our work outperforms the joint optimization frameworks \ac{nerf}$\text{-}\text{-}$ and SinNeRF on existing datasets. By using our Gaussian Fourier feature-based pose \ac{mlp}, we achieve a better camera pose estimation, which consequently enables an improved estimation of the intrinsic camera parameters and higher-quality \ac{nvs} results. In summary, \approach allows the estimation of both diverse intrinsic and extrinsic camera parameters in joint optimization with \ac{nerf}.

\bibliographystyle{IEEEtran} % We choose the "IEEEtran" reference style
\bibliography{IEEEabrv, paper} % Entries are in the refs.bib file

\newpage\null\thispagestyle{empty}\newpage
\beginsupplement

%%%%%%%%% BODY TEXT
\newpage\null\thispagestyle{empty}
\begin{minipage}[t]{\textwidth}
\centering
\textbf{\huge{Supplementary Material}}
\end{minipage}

\newpage\null\thispagestyle{empty}\newpage\null\thispagestyle{empty}

\section{Camera Parameter Experiments}
\approach aims to improve the extrinsic camera parameter estimation using Gaussian Fourier features and to predict differing intrinsic camera parameters if they are given in a dataset. We first show a deeper analysis of extrinsic camera parameter estimation. Secondly, we show additional results for the camera intrinsic parameter estimation. We further analyze novel view synthesis. Lastly, we provide details about the focal length and example images of our \datasetname dataset.

\subsection{Extrinsic Camera Parameters}

For the extrinsic camera parameters we conducted additional experiments. In these we compare positional encoding and Gaussian Fourier features as well as different embedding sizes for our pose \ac{mlp}. Furthermore, we analyze mirror poses, investigate the COLMAP initialization, the influence of the \ac{ssim} loss and conduct another breaking point analysis.

\begin{table}[t!]
    \centering

   \begin{center}
    \resizebox{\columnwidth}{!}{ 
    \begin{tabular}{l|cc|cc} \hline \hline
       \multirow{2}{*}{Scene} & \multicolumn{2}{c|}{Pos. Enc.} & \multicolumn{2}{c}{Gaussian F.F.} \\
        & Rot. Err. & Trans. Err. & Rot. Err. & Trans. Err. \\ \hline 
       Fern     & 36.18 & 0.035 & \textbf{23.28} & \textbf{0.026}  \\
       Flower   & 39.64 & \textbf{0.025} & \textbf{5.36} & 0.027  \\
       Fortress & 38.05 & \textbf{0.033} & \textbf{27.47} & 0.047  \\
       Horns    & 46.51 & \textbf{0.051} & \textbf{4.90} & \textbf{0.051}  \\
       Leaves   & 42.55 & \textbf{0.020} & \textbf{4.03} & 0.022 \\
       Orchids  & \textbf{22.48} & \textbf{0.025} & 23.73 & \textbf{0.025}  \\
       Room     & 64.49 & 0.086 & \textbf{41.03} & \textbf{0.069}  \\
       T-Rex    & \textbf{77.77} & 0.065 & 85.50 & \textbf{0.052}\\ \hline
       Mean     & 45.95 & 0.042 & \textbf{26.91} & \textbf{0.039}  \\ \hline \hline
    \end{tabular}
    }
   \end{center}
    \caption{\textbf{Quantitative comparison between positional encoding and Gaussian Fourier features on \acs{llff}.} The results are computed after 500 epochs. The high error values are the result of ten random seeds for each scene.}
    \label{tab:pos_encVSgaussian}
\end{table}

\begin{table}[t!]
    \centering
   \begin{center}
    \resizebox{\columnwidth}{!}{ 
    \begin{tabular}{l|cc|cc|cc} \hline \hline
       \multirow{2}{*}{Scene} & \multicolumn{2}{c|}{\acs{psnr}} & \multicolumn{2}{c|}{Rot. Err.} & \multicolumn{2}{c}{Trans. Err.} \\
        & 64 & 256  & 64 & 256  & 64 & 256 \\ \hline 
       Fern     & 21.63  & 22.75 & 1.59 & 0.94 & 0.009 & 0.004\\
       Flower   & 24.66 & 25.95 & 3.98 & 1.59  & 0.023 & 0.017 \\
       Fortress & 27.24 & 28.39 & 1.30 & 1.32 & 0.014 & 0.015\\
       Horns    & 22.46 & 25.20 & 1.05 & 1.45 & 0.014  & 0.005\\
       Leaves   & 17.24 & 19.00 & 4.26 & 4.03 & 0.012 & 0.016\\
       Orchids  & 17.39 & 16.87 & 2.86 & 4.12 & 0.005 & 0.024 \\
       Room     & 27.04 & 27.54 & 0.97  & 1.15 & 0.009 & 0.036\\
       T-Rex    & 23.04 & 24.25 & 3.36 & 3.85 & 0.012 & 0.009 \\ \hline
       Mean     & 22.59 & \textbf{23.74}  & 2.42 & \textbf{2.31}  & \textbf{0.012}  & 0.016\\ \hline \hline
    \end{tabular}
    }
   \end{center}
    \caption{\textbf{Quantitative comparison between two different embedding sizes for the pose \acs{mlp}, i.e. 64 and 256 on \acs{llff}.} The embedding size of 256 leads to an higher \acs{psnr} and lower rotation error.}
    \label{tab:64vs256embeddingsize}
\end{table}
\begin{figure}[t!]
\centering
    \subfloat{\includegraphics[width=\columnwidth]{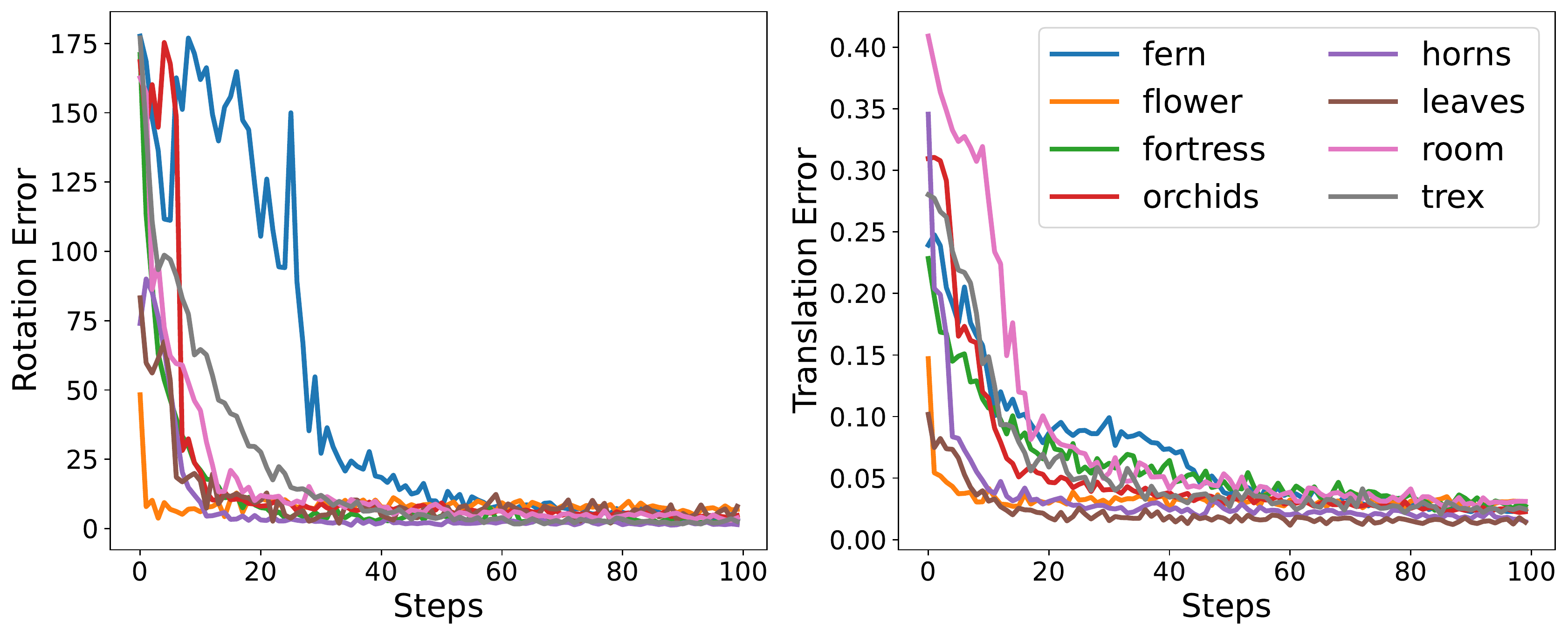}}\hskip1ex
    \subfloat{\includegraphics[width=\columnwidth]{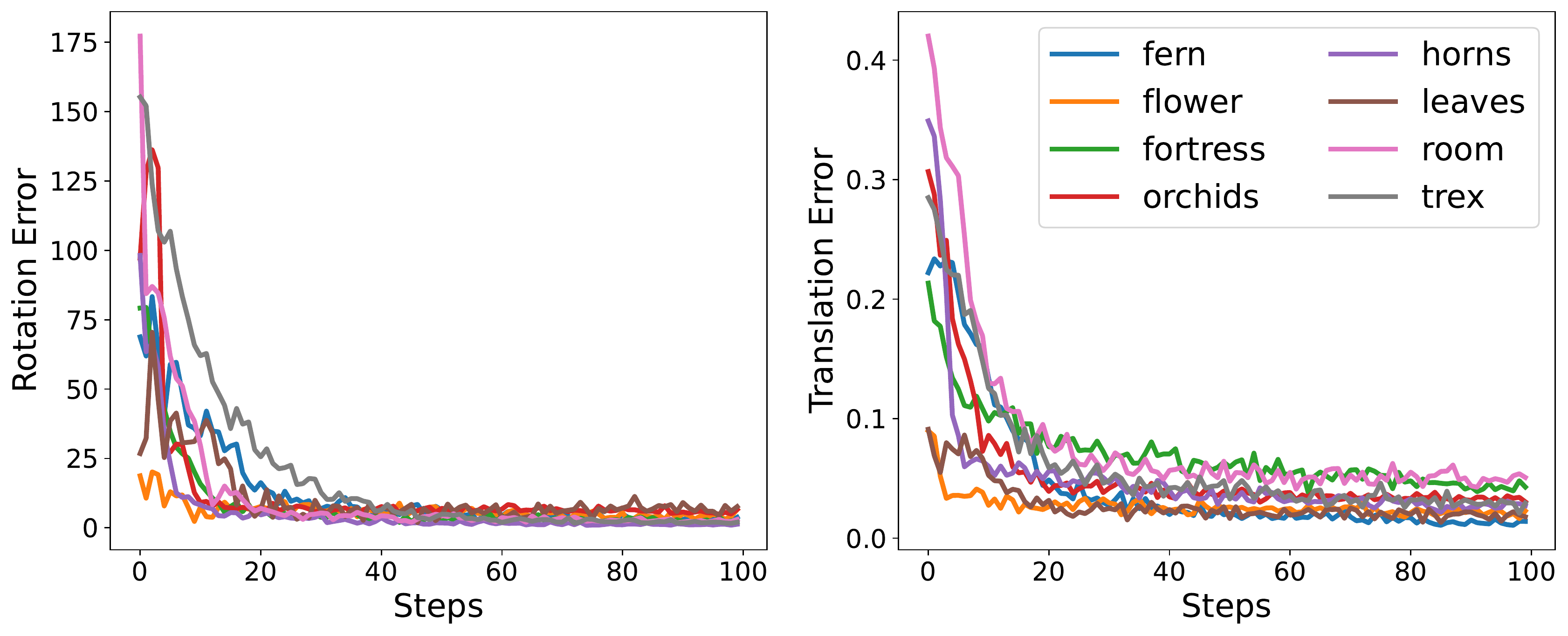}}
    \caption{\textbf{Rotation error and translation error for the first 100 epochs between two different embedding sizes i.e. 64 (top) and 256 (bottom). } The embedding size 256 shows less outliers compared to an embedding size of 64.}
\label{fig:embeddingsizePlot}
\end{figure}

\subsubsection{Positional Encoding}
Our pose \ac{mlp} uses Gaussian Fourier features while another approach is the use of positional encoding~\cite{lin_barf_2021,truong_sparf_2022}. Therefore, we test positional encoding in our pose \ac{mlp}. We train our pose \ac{mlp} on each scene of the \ac{llff} dataset~\cite{mildenhall_local_2019}. For training we use ten random seeds.  In Table~\ref{tab:pos_encVSgaussian} we compare the results of Gaussian Fourier features and positional encoding after 500 epochs. Gaussian Fourier features averagely outperform positional encoding in mean rotation and translation error over the ten runs. The high mean errors occur due to mirror poses in the ten runs. We also include  the experiments with a mirror pose in our calculation to show that mirror poses are less likely to occur when using Gaussian Fourier features.

\subsubsection{Embedding Size}
In addition to testing positional encoding, we investigate the influence of different embedding sizes for the pose \ac{mlp}, see Table~\ref{tab:64vs256embeddingsize}. We apply Gaussian Fourier feature mapping and use the embedding sizes 64 and 256 respectively. As visualized in Fig.~\ref{fig:embeddingsizePlot} the embedding size of 64 shows more outliers for the rotation convergence, for example in the fern and orchids scene, and for the translation convergence in the room scene. Based on the results showing more stability of 256, we select the embedding size of 256 for our pose \ac{mlp}.

\begin{table*}
    \centering
       \resizebox{\textwidth}{!}{ 
    \begin{tabular}{|c|P{40pt}P{40pt}|P{40pt}P{40pt}|P{40pt}P{40pt}|P{40pt}P{40pt}|P{40pt}P{40pt}|} \hline
    Skip/Scene & \multicolumn{2}{c|}{Fireplug} &  \multicolumn{2}{c|}{Stormtrooper}  &  \multicolumn{2}{c|}{T1}  & \multicolumn{2}{c|}{Bike} & \multicolumn{2}{c|}{Brick House} \\ \hline
     & \scriptsize{\acs{nerf}$\text{-}\text{-}$}  & \scriptsize{OURS}  & \scriptsize{\acs{nerf}$\text{-}\text{-}$}  & \scriptsize{OURS} & \scriptsize{\acs{nerf}$\text{-}\text{-}$}  & \scriptsize{OURS} & \scriptsize{\acs{nerf}$\text{-}\text{-}$}  & \scriptsize{OURS} & \scriptsize{\acs{nerf}$\text{-}\text{-}$}  & \scriptsize{OURS} \\ 
    2 & \green & \green & \green & \green  & \green & \green & \red & \green & \green & \green  \\
    3 & \green & \green & \green & \green  & \green & \green & \red & \red & \green & \green  \\
    4 & \green & \green & \green & \red & \red & \green & \red & \green & \green & \green  \\
    5 & \green & \red & \green & \red  & \red & \green & \red & \red & \green & \green  \\ \hline

    \end{tabular}
    }
    \caption{\textbf{Breaking point analysis of our Gaussian Fourier feature-based Pose \acs{mlp} vs. \acs{nerf}$\text{-}\text{-}$ on \datasetname.} We used every second, third, fourth or fifth image during training. A rotation error below  $20^{\circ}$ is considered as success (green). We slightly outperform \acs{nerf}$\text{-}\text{-}$ as we succeed on 15 scenes while \acs{nerf}$\text{-}\text{-}$ only succeeds on 14 scenes.\label{tab:breaking_ana_iff}}
\end{table*}

\subsubsection{Mirror Poses}

Wang et al.~\cite{wang_nerf_2021} mentioned the occurrence of mirror poses when using COLMAP. In their experiments on the \ac{bleff} dataset, COLMAP shows an error close to $180^{\circ}$ for the roundtable scene. For the parameter-based or \ac{mlp}-based pose estimation similar errors can appear. This can occur due to an ambiguous pose estimation caused by several local minima in the loss function~\cite{schweighofer_robust_2006,oberkampf_iterative_1996}.

\begin{figure}[t!]
  \centering
    \includegraphics[width=\columnwidth]{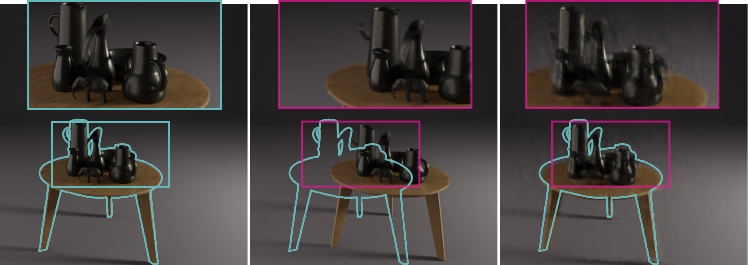}
  \caption{\textbf{Comparison of the same scene of \ac{bleff} and same frame at different rotation error levels.} We highlight the original shift from the ground truth scene (left) with a blue outline around the table. In the pink rectangles we show the same cutout from the same coordinates in the image. 
  % The actual ground truth scene (left) and rotation error near 180 degrees and a \ac{psnr} value of $47.13$ (middle) against a novel view with error at 28 degrees with a \ac{psnr} value of $39.69$ (right). 
  }
  \label{fig:roundtable_error}
\end{figure}

Nevertheless, the \ac{nerf} does not fail to learn and generate accurate novel views. Despite high image quality, a clear shift of the objects in the scene is visible. Especially for the roundtable scene, low rotation errors lead to a better result in the rotation comparison. Still, the visual novel view quality is poor despite high \ac{ssim} and \ac{psnr} values. An example of this can be seen in Fig.~\ref{fig:roundtable_error}. The picture in the center of Fig.~\ref{fig:roundtable_error} shows a mirror pose that occurred during training with a train \ac{psnr} of $47.13$ and an average rotation error above $150^{\circ}$. However, the generated image is of high quality compared to the image on the right. There, the scene has a comparatively low average rotation error of below $30^{\circ}$, but only a train \ac{psnr} of $39.69$. Although the rotation error in the right image is $<30^{\circ}$, the \ac{psnr} value for the middle figure is much higher. Nevertheless, the outline of the ground truth scene shows a highly incorrect camera pose in the center image. Thus, both metric types, \ac{nvs} quality metrics and \ac{ate}, are decisive for joint optimization approaches.

\begin{figure}[t!]
\centering
    \subfloat{\includegraphics[width=\columnwidth]{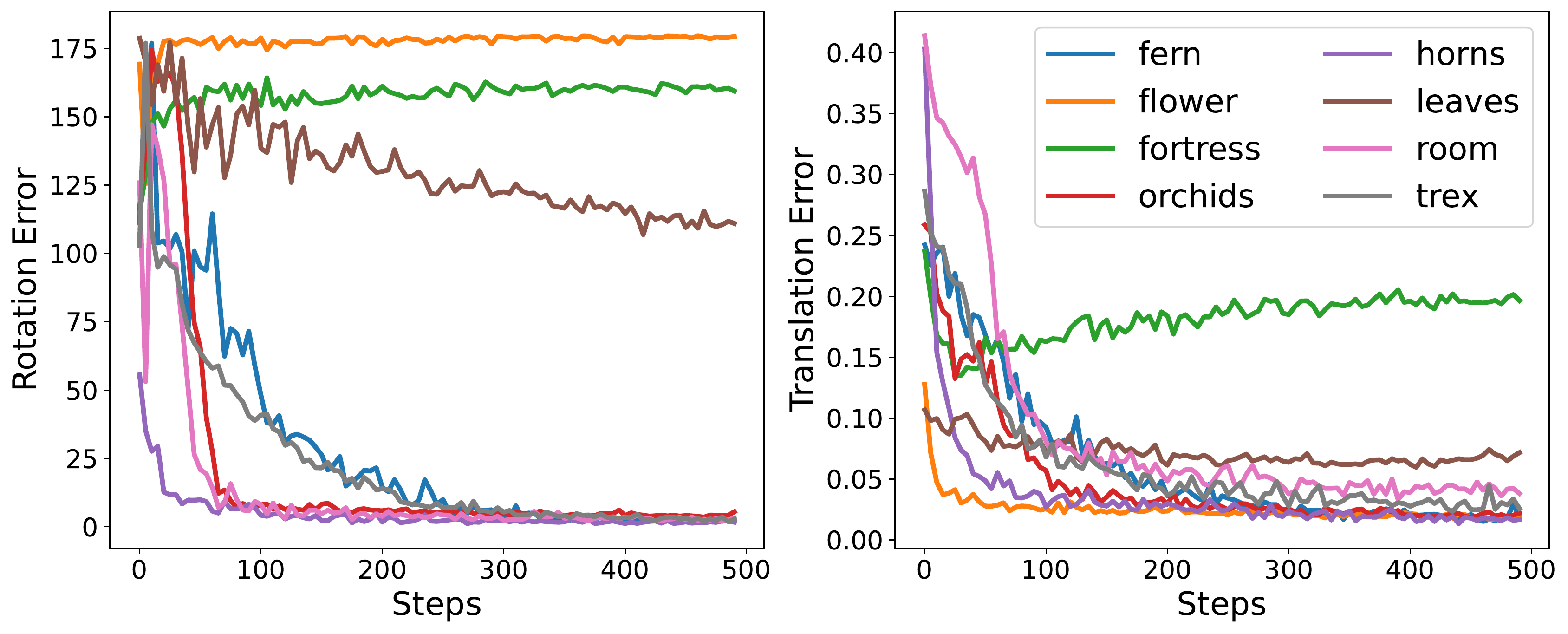}}
    
    \subfloat{\includegraphics[width=\columnwidth]{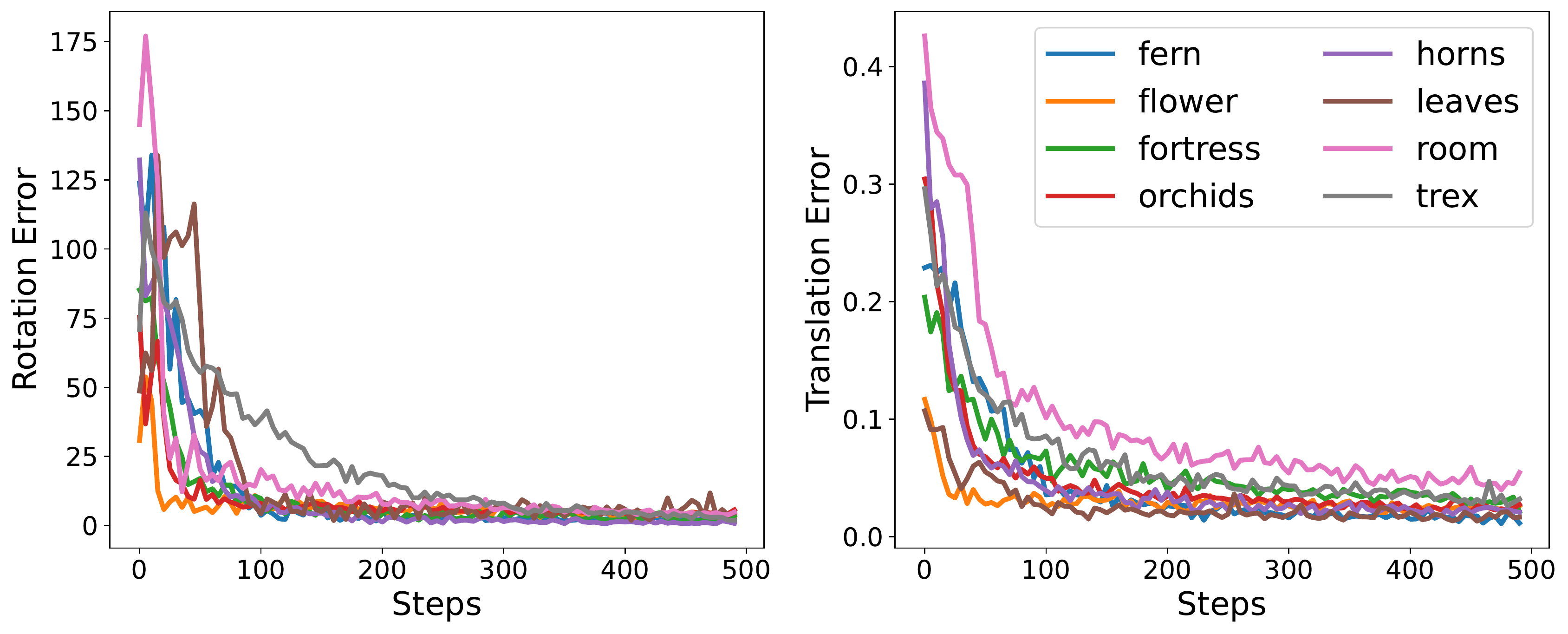}}
    \caption{\textbf{Comparison of ATE statistics on all \acs{llff} scenes.} We show the comparison without \acs{ssim} loss (top) and with \acs{ssim} loss (bottom) for 500 epochs.}
\label{fig:ssimcomp}
\end{figure}

\begin{table*}[t!]
    \centering

   \begin{center}
    \resizebox{\textwidth}{!}{ 
    \begin{tabular}{l|cccc|cccc|cccc} \hline \hline
       \multirow{2}{*}{Scene} & \multicolumn{4}{c|}{Rot. Err.} & \multicolumn{4}{c|}{Trans. Err.} & \multicolumn{4}{c}{Focal. Err.} \\
        & \acs{nerf}$\text{-}\text{-}$& SiNeRF & OURS & OURS$^\text{(COLMAP)}$ & \acs{nerf}$\text{-}\text{-}$& SiNeRF & OURS & OURS$^\text{(COLMAP)}$ & \acs{nerf}$\text{-}\text{-}$& SiNeRF & OURS & OURS$^\text{(COLMAP)}$\\ \hline 
       Fern     &1.78 & 1.17 & 1.34 & \textbf{0.41} & 0.029  &  0.006 & 0.006 & \textbf{0.001} & 153.5 & 112.7 & 161.6 & \textbf{1.76}\\
       Flower   & 4.84 & 1.38 & 0.89 & \textbf{0.39} & 0.016 & 0.007 & 0.007 & \textbf{0.002} & \textbf{13.2} & 80.4 & 53.6 & \textbf{18.63}\\
       Fortress & 1.36 & 2.02 & 0.91 & \textbf{0.15} & 0.025 & 0.048 & 0.006 & \textbf{0.002} & 144.1 & 59.7 & 129.3 & \textbf{1.05} \\
       Horns    & 5.55 & 83.34 & 1.89 & \textbf{0.41} & 0.044 & 0.133 & 0.014 & \textbf{0.002} & 156.2 &282.5 & 92.5 & \textbf{25.8} \\
       Leaves   & 3.90 & 14.46 & \textbf{2.65} & 5.04 & 0.016 & 0.10 & 0.005 & \textbf{0.004} & 59.0 & 18.8 &  24.4 & \textbf{10.3} \\
       Orchids  & 4.96 & 3.97 & 2.75 &\textbf{0.31} & 0.051 & 0.014 & 0.009 & \textbf{0.004} & 199.3 & 155.7 & 165.9 & \textbf{11.96} \\
       Room     & 2.77 & 4.92 & 1.33 & \textbf{0.14} & 0.030 & 0.022 & 0.006 & \textbf{0.001} & 331.8 &313.6 & \textbf{102.7} & \textbf{6.06} \\
       T-Rex    & 4.67 & 7.19 & \textbf{3.90} & \textbf{0.14} & 0.036 & 0.027 & 0.008 & \textbf{0.002} & 89.3 & 38.3 & 100.4 & \textbf{7.54} \\ \hline
       Mean     & 3.73 & 14.81 & \textbf{2.10} & \textbf{0.83} & 0.031 & 0.033 & 0.008 & \textbf{0.002} & 143.3 &  155.2 & 103.7 & \textbf{10.4} \\ \hline \hline
    \end{tabular}
    }
   \end{center}
    \caption{\textbf{Quantitative comparison between \approach (OURS),  \approach initialized with COLMAP (OURS$^{COLMAP}$), \acs{nerf}$\text{-}\text{-}$~\cite{wang_nerf_2021} and SiNeRF~\cite{xia_sinerf_2022} on \acs{llff}.} For SiNeRF we retrained the approach with a layer dimension of 128 to ensure comparability with our approach and \ac{nerf}$\text{-}\text{-}$. We report the camera extrinsic parameter errors, in detail translation and rotation error. We tested ten random seeds for SiNeRF; for \ac{nerf}$\text{-}\text{-}$ we report the values from Wang et. al~\cite{wang_nerf_2021}.}
    \label{tab:extrinsic_llff_colmap}
\end{table*}

\begin{table*}[t]
   \begin{center}
       \resizebox{\textwidth}{!}{ 
    \begin{tabular}{l|cccc|cccc|cccc} \hline \hline
       \multirow{2}{*}{Scene} & \multicolumn{4}{c|}{PSNR$\uparrow$} & \multicolumn{4}{c|}{SSIM$\uparrow$} & \multicolumn{4}{c}{LPIPS$\downarrow$} \\  % \cmidrule(r){2.5-4.5}
        & \ac{nerf} & \ac{nerf}$\text{-}\text{-}$ & OURS & OURS$^\text{(COLMAP)}$ & \ac{nerf} & \ac{nerf}$\text{-}\text{-}$ &  OURS & OURS$^\text{(COLMAP)}$ &  \ac{nerf} & \ac{nerf}$\text{-}\text{-}$ & OURS & OURS$^\text{(COLMAP)}$\\ \hline
       Fern     & \textbf{22.22} & 21.67 & 21.82 & 22.19 & \textbf{0.64} & 0.61 & 0.62 & \textbf{0.64} & 0.47 & 0.50 & 0.49 & 0.41 \\
       Flower   & 25.25 & 25.34 & 25.39 & \textbf{25.78} & 0.71 & 0.71 & 0.72 & \textbf{0.73} & 0.36 & 0.37 & 0.37 & \textbf{0.35}\\
       Fortress & 27.60 & 26.20 & 27.28 & \textbf{28.13} & 0.73 & 0.63 & 0.70 & \textbf{0.74} & 0.38 & 0.49 & 0.41 & \textbf{0.37} \\
       Horns   & 24.25 & 22.53 & 24.00 & \textbf{24.45} & 0.68 & 0.61 & 0.67 & \textbf{0.68} & \textbf{0.44} & 0.50 & 0.45 & \textbf{0.44} \\
       Leaves   & 18.81 & 18.88 & \textbf{18.97} & 16.65 & 0.52 & \textbf{0.53} & 0.50 & 0.43 & \textbf{0.47} & \textbf{0.47} & 0.49 & 0.53\\
       Orchids  & 19.09 & 16.73 & 17.41 & \textbf{19.30} & 0.51 & 0.39 & 0.43 & \textbf{0.52} & \textbf{0.46} & 0.55 & 0.52 & \textbf{0.46} \\
       Room     & \textbf{27.77} & 25.84 & 27.12 & 27.72 & \textbf{0.87} & 0.84 & 0.82 & \textbf{0.87} & 0.40 & 0.44 & 0.43 & \textbf{0.41} \\
       T-Rex    & 23.19 & 22.67 & 22.92 & \textbf{23.20} & \textbf{0.74} & 0.72 & \textbf{0.74} & \textbf{0.74} & \textbf{0.41} & 0.44 & 0.43 & \textbf{0.41} \\ \hline
       Mean     & 23.52 & 22.48 & 23.14 & \textbf{23.70} & 0.68 & 0.63 & 0.66 & \textbf{0.74} & 0.42 & 0.47 & 0.44 & \textbf{0.41} \\ \hline \hline
    \end{tabular}
    }
   \end{center}
     \caption{\textbf{Quantitative comparison between \approach (OURS),  \approach with COLMAP initialization (OURS$^{COLMAP}$), \ac{nerf}$\text{-}\text{-}$~\cite{wang_nerf_2021} and COLMAP-based \ac{nerf}~\cite{wang_nerf_2021} on \ac{llff}}. We follow the results of Wang et. al~\cite{wang_nerf_2021} for COLMAP-based \ac{nerf}. Thus, the reported \ac{nvs} quality is overall lower compared to vanilla \ac{nerf} as the layer size is 128 instead of 256. We report \ac{psnr}, \ac{ssim} and \acs{lpips}. When using COLMAP initialization we outperform COLMAP-based \acs{nerf} in \acs{nvs} quality.}
    \label{tab:benchmark_llff_colmap_init}
\end{table*}

\subsubsection{SSIM Loss}
To test whether the additional \ac{ssim} loss supports the pose estimation, we train on each scene of \ac{llff} for 500 epochs with fixed seeds. As can be observed in Fig.~\ref{fig:ssimcomp} the altered training objective leads to a convergence in scenes where a mirror pose would otherwise occur. The convergence speed for the rotation and translation error also increases. Since the \ac{psnr} is lower when \ac{ssim} loss is used in the whole training process, we apply it only in the first 500 epochs.

\subsubsection{Breaking Point Analysis on \datasetname}

Our main paper includes a breaking point analysis on \ac{llff}, where we outperform  \ac{nerf}$\text{-}\text{-}$. We also run this comparison on \datasetname. A rotation error lower than $20^{\circ}$ is considered as success. For the training, we used every second, third, fourth and fifth image. As shown in Table~\ref{tab:breaking_ana_iff}, we succeed in one more scene than \ac{nerf}$\text{-}\text{-}$. This allows the conclusion, that differing intrinsic camera parameters make the estimation of camera extrinsic parameters more challenging.

\begin{table*}[t!]
   \begin{center}
       \resizebox{\textwidth}{!}{ 
    \begin{tabular}{l|ccc|ccc|ccc|ccc|ccc} \hline \hline
       \multirow{2}{*}{Scene} & \multicolumn{3}{c|}{PSNR$\uparrow$} & \multicolumn{3}{c|}{SSIM$\uparrow$} & \multicolumn{3}{c|}{LPIPS$\downarrow$} & \multicolumn{3}{c|}{Rot. Err.} & \multicolumn{3}{c}{Trans. Err.} \\  % \cmidrule(r){2.5-4.5}
        & \ac{nerf}$\text{-}\text{-}$ & SiNeRF & OURS & \ac{nerf}$\text{-}\text{-}$ &  SiNeRF & OURS &  \ac{nerf}$\text{-}\text{-}$ & SiNeRF & OURS  & \ac{nerf}$\text{-}\text{-}$& SiNeRF & OURS & \ac{nerf}$\text{-}\text{-}$& SiNeRF & OURS  \\ \hline
       Fern & 22.15 & 22.48 &  \textbf{22.75} & 0.65 & 0.67 &  \textbf{0.69} & 0.46 & 0.44 &  \textbf{0.41} & 1.57 & 0.74 & 0.94 & 0.008 & 0.004 & 0.004\\
       Flower & 26.61 & \textbf{27.23} & 25.95 & 0.77 & \textbf{0.80} & 0.74 & 0.30 & 0.30 & 0.33 & 3.21 & 0.51 & 2.3 & 0.012 & 0.008 & 0.022\\
       Fortress & 25.60 & 27.47 &  \textbf{28.39} & 0.60 & 0.72  &  \textbf{0.77} & 0.54 & 0.39 &  \textbf{0.31} & 2.41 & 1.77 & 1.32 & 0.060 & 0.041 & 0.015\\
       Horns & 23.17 & 24.14 &  \textbf{25.20} & 0.64 & 0.68 &  \textbf{0.73} & 0.51 & 0.43 &  \textbf{0.37} & 3.04 & 2.66 & 1.45 & 0.015 & 0.022 & 0.005\\
       Leaves &  \textbf{19.74} & 19.15 & 19.00 &  \textbf{0.61} & 0.57 & 0.55 &  \textbf{0.39} &  \textbf{0.39} & 0.43  & 6.78 & 8.76 & 4.03 & 0.006 & 0.008 & 0.016\\
       Orchids & 15.86 &  \textbf{16.92} & 16.87 & 0.35 &  \textbf{0.41} &  \textbf{0.41} & 0.55 & 0.53 &  \textbf{0.50} & 5.46 & 3.24 & 4.19 & 0.022 & 0.013 & 0.024\\
       Room  & 25.68 & 26.10 &  \textbf{27.54} & 0.84 & 0.84 &  \textbf{0.87} & 0.41 & 0.43 &  \textbf{0.37} & 3.75 & 2.08 & 1.15 & 0.021 & 0.021 & 0.036\\
       T-Rex & 23.38 & \textbf{24.94} & 24.25 & 0.76 & \textbf{0.82} & 0.79 & 0.39 & 0.36 & 0.36 & 6.34 & 0.86 & 3.36 & 0.015 & 0.005 & 0.009\\ \hline
       Mean & 22.77 & 23.55 & \textbf{23.74} & 0.65 & \textbf{0.69} & \textbf{0.69} & 0.44 & 0.41 & \textbf{0.39} & 4.07 & 2.58 & \textbf{2.32} & 0.019 & \textbf{0.015} & 0.016 \\ \hline \hline
    \end{tabular}
    }
   \end{center}
     \caption{\textbf{Quantitative comparison between \approach (OURS), \ac{nerf}$\text{-}\text{-}$~\cite{wang_nerf_2021} and SiNeRF~\cite{xia_sinerf_2022} on  \ac{llff}, with a layer dimension of 256 for all \ac{nerf} frameworks.} We follow Xia et. al~\cite{xia_sinerf_2022} for the results of \ac{nerf}$\text{-}\text{-}$ with a layer dimension of 256. We report \ac{psnr}, \ac{ssim}, \acs{lpips} and the extrinsic camera error, in detail translation and rotation error. \approach outperforms \ac{nerf}$\text{-}\text{-}$~\cite{wang_nerf_2021} and SiNeRF in \acs{psnr} and \acs{lpips}.}
    \label{tab:benchmark_llff_256}
\end{table*}

\subsubsection{COLMAP Initialization for Camera Parameter Estimation}
\approach per se learns the camera parameters from scratch,{so without prior initialization}. 
Besides learning the camera parameters without initializing the pose \ac{mlp}, COLMAP pre-processed poses can be used for its initialization. Research on $360^{\circ}$ pose estimation and joint \ac{nvs} optimization initializes the pose network either directly with COLMAP or refines from noisy poses also calculated from COLMAP poses~\cite{lin_barf_2021}.{For ablation studies,} we initialize the pose \ac{mlp} with COLMAP poses and then train the joint optimization. The results are denoted in Table~\ref{tab:extrinsic_llff_colmap} for \ac{llff} and in Table~\ref{tab:extrinsic_iff_colmap} for \datasetname. We provide results for joint optimization approaches and for the COLMAP baseline on both dataset. Again, the results show that our improved pose prediction reduces the focal length error and improves \ac{nvs}. Also, on \datasetname with diverse intrinsic and extrinsic parameters our approach outperforms COLMAP-based \ac{nerf}, \ac{nerf} in the joint optimization.
% Besides training the network from scratch we also tested how our approach performs when using COLMAP initialization. 

\begin{table}[t!]
    \centering

   \begin{center}
    % \resizebox{\columnwidth}{!}{ 
    \begin{tabular}{l|ccc} \hline \hline
       \multirow{2}{*}{Scene} & \multicolumn{3}{c}{Focal. Err.} \\
        & \ac{nerf}$\text{-}\text{-}$ & SiNeRF & OURS \\ \hline 
       Fern     & 153.5 & \textbf{112.7} & 161.6\\
       Flower   & \textbf{13.2} & 80.4 & 53.6\\
       Fortress & 144.1 & \textbf{59.7} & 129.3 \\
       Horns    & 156.2 & 282.5 & \textbf{92.5} \\
       Leaves   & 59.0 & 198.8 &  \textbf{24.4} \\
       Orchids  & 199.3 & \textbf{155.7} & 165.9 \\
       Room     & 331.8 & 313.6 & \textbf{102.7}\\
       T-Rex    & 89.3 & \textbf{38.3} & 100.4\\ \hline
       Mean     & 143.3 &  155.2 & \textbf{103.7} \\ \hline \hline
    \end{tabular}
   %  }
   \end{center}
    \caption{\textbf{Quantitative comparison between \approach (OURS), \ac{nerf}$\text{-}\text{-}$~\cite{wang_nerf_2021} and SiNeRF~\cite{xia_sinerf_2022} with a \ac{nerf} layer dimension of 128 on \ac{llff}.} We report the focal pixel error. Our approach outperforms \ac{nerf}$\text{-}\text{-}$~\cite{wang_nerf_2021} and SiNeRF~\cite{xia_sinerf_2022}.}
    \label{tab:extrinsic_and_intrinsics_llff}
\end{table}

\begin{table}[t!]
    \centering
    
       \resizebox{\columnwidth}{!}{ 
    \begin{tabular}{l|l} \hline \hline
        Scene & Focal Lengths \\ \hline
        T1 &{$16\times740.90$, $14\times1673.73$, $1\times980.71$}  \\ 
        Fireplug &{$21\times3531.73$, $2\times2663.47$, $8\times683.74$} \\ 
        Bike &{$16\times3022.43$, $15\times756.16$} \\ 
        Stormtrooper &{$29x 1573.18, 1x 1380.96, 1x 585.26$} \\ 
        Brick House &{$10\times3000.51$, $2\times3164.69.86$, $10\times1483.85$, $9\times746.25$} \\ \hline \hline
    \end{tabular}
    }
    
    \caption{\textbf{Focal lengths of the individual scenes from \datasetname.} The camera parameters were estimated using COLMAP.}
    \label{tab:ibleff_license}
\end{table}

\begin{table*}[t!]
    \centering

   \begin{center}
    \resizebox{\textwidth}{!}{     
    \begin{tabular}{l|cccc|cccc|ccc|ccc|ccc} \hline \hline
        \multirow{2}{*}{Scene} & \multicolumn{4}{c|}{\acs{psnr}} & \multicolumn{4}{c|}{\acs{ssim}} & \multicolumn{3}{c|}{Focal Err.} & \multicolumn{3}{c|}{Rot Err.} & \multicolumn{3}{c}{Trans Err.}\\
        &{\ac{nerf}} & \ac{nerf}$\text{-}\text{-}$ & \ac{nerf}$\text{-}\text{-}$ + I & I+GF &{\ac{nerf}} & \ac{nerf}$\text{-}\text{-}$ & \ac{nerf}$\text{-}\text{-}$ + I & I+GF  & \ac{nerf}$\text{-}\text{-}$ & I & I+GF & \ac{nerf}$\text{-}\text{-}$ & I & I+GF & \ac{nerf}$\text{-}\text{-}$ & I & I+GF\\ \hline  
        T1 &{25.91} & 27.26 & 27.61 & \textbf{28.86} &{0.83} & 0.86 & 0.85 & \textbf{0.89} & 207.96 & 76.11 & 50.41 & 2.31 & 1.15 & 0.34 & 0.174 & 0.022 & \textbf{0.010}\\ 
        Brick House &{25.44} & 24.94 & 27.85 & 28.57 &{0.74} & 0.75 & 0.77 & \textbf{0.79} & 170.07 & 51.63 & 28.85 & \textbf{2.45} & 5.29 & 3.09 & 0.148 & 0.046 & \textbf{0.034} \\
        Fireplug &{22.60} & 22.52 & 22.95 & \textbf{25.15} &{0.65} & 0.61 & 0.63 & \textbf{0.71} & 267.89 & 123.24 & \textbf{92.14} & \textbf{5.45} & 5.61 & 5.51 & 0.006 & \textbf{0.001} & 0.007 \\
        Bike &{25.81} & 15.94 & 22.38 & \textbf{26.53} &{0.84} & 0.27 & 0.71 & \textbf{0.85} & 326.48 & 207.72 & \textbf{43.61} & 7.50 & 4.71 & \textbf{3.26} & 0.080 & 0.031 & 0.001 \\
        Stormtropper &{25.00} & 30.00 & 34.61 & \textbf{36.99} &{0.90} & 0.94 & 0.98 & \textbf{0.99} & 121.47 & 74.83 & \textbf{34.69} & \textbf{7.40} & \textbf{7.40} & 11.54 & \textbf{0.896} & \textbf{0.896} & 1.224 \\
        Mean &{24.95} & 24.16 & 27.57 & \textbf{29.50} & 0.79 & 0.69 & 0.81 & \textbf{0.86} & 208.44 & 101.61 & \textbf{48.06} & 5.41 & 4.85 & \textbf{4.53} & 0.263 & \textbf{0.200} & 0.232 \\ \hline \hline
    
    \end{tabular}
   }
   \end{center}
    \caption{\textbf{Quantitative comparison between{COLMAP-based \ac{nerf}}, \ac{nerf}$\text{-}\text{-}$~\cite{wang_nerf_2021}, \ac{nerf}$\text{-}\text{-}$ and our improved intrinsic estimation (I) and \approach (I+GF) on \datasetname with COLMAP initialization.} We report \acs{psnr}, \acs{ssim}, rotation error, translation error and focal length error.}
    \label{tab:extrinsic_iff_colmap}
\end{table*}

\subsection{Intrinsic Camera Parameters}

In addition to the extrinsic camera parameter estimation, we compare our intrinsic camera parameter estimation against \ac{nerf}$\text{-}\text{-}$ and SiNeRF on \ac{llff}. As denoted in Table~\ref{tab:extrinsic_and_intrinsics_llff}, we outperform both approaches in the mean focal length prediction. While we clearly outperform \ac{nerf}$\text{-}\text{-}$ in the individual values, SiNeRF shows comparable results. However, SiNeRF produces a significant error in the room scene. Thus, their mean focal length error suffers. When excluding the room scene from the mean value calculation, we still outperform SiNeRF, with a focal pixel error of 103.9 compared to 132.6.

Moreover, the COLMAP initialization tested on \ac{llff} shows also a lower focal length error.  As denoted in Table~\ref{tab:extrinsic_llff_colmap},   a reduced pose error leads to a lower focal length error.

\subsection{A Higher \acs{nerf} Layer Dimension and its Influence on Camera Parameters}

\approach is off-the-shelf comparable with \ac{nerf}$\text{-}\text{-}$, as we use a layer dimension of 128. SiNeRF was originally trained with a layer dimension of 256. When applying this layer dimension to our \ac{nerf}, the results for the intrinsic and extrinsic camera parameters change due to the joint optimization. For the focal length we report a mean error of 72.12. With a layer dimension of 128 we reported a mean focal length error of 103.7, see Table~\ref{tab:extrinsic_and_intrinsics_llff}. The extrinsic camera parameter results are denoted in Table~\ref{tab:benchmark_llff_256}. \approach outperforms the other joint optimization approaches in rotation error and translation error when using a layer dimension of 256.

\section{Novel View Synthesis} 

\subsection{COLMAP Initialization for NVS Quality}

Initializing the pose \ac{mlp} with COLMAP shows an improved prediction of the intrinsic and extrinsic camera parameters, see Table~\ref{tab:extrinsic_llff_colmap}{and Table~\ref{tab:extrinsic_iff_colmap}}. Our further experiments{on \ac{llff} and \datasetname} demonstrate that a better pose estimation leads to an improved \ac{nvs}. Table~\ref{tab:benchmark_llff_colmap_init} shows the \ac{psnr}, \ac{ssim} and \ac{lpips} results for all scenes on \ac{llff}. The \ac{nvs} quality is higher in the joint optimization than when purely training on COLMAP poses. These findings corrospond with other related works~\cite{lin_barf_2021,truong_sparf_2022}. % They already show that joint optimization of camera poses and \ac{nerf} lead to increased \ac{nvs} quality.  

\subsection{Higher Layer Dimension for NVS Quality}
In our ablation study we compare \approach with the original SiNeRF and their adapted \ac{nerf}$\text{-}\text{-}$ with a layer dimension of 256, see Table~\ref{tab:benchmark_llff_256}. \approach outperforms both other joint optimization  approaches. We achieve better results in \ac{psnr} and \ac{lpips}. For \ac{ssim} we perform equally compared to SiNeRF but outperform \ac{nerf}$\text{-}\text{-}$.

\begin{figure*}[t!]
  \centering
    \subfloat[Brick House]{
        \includegraphics[height=2cm]{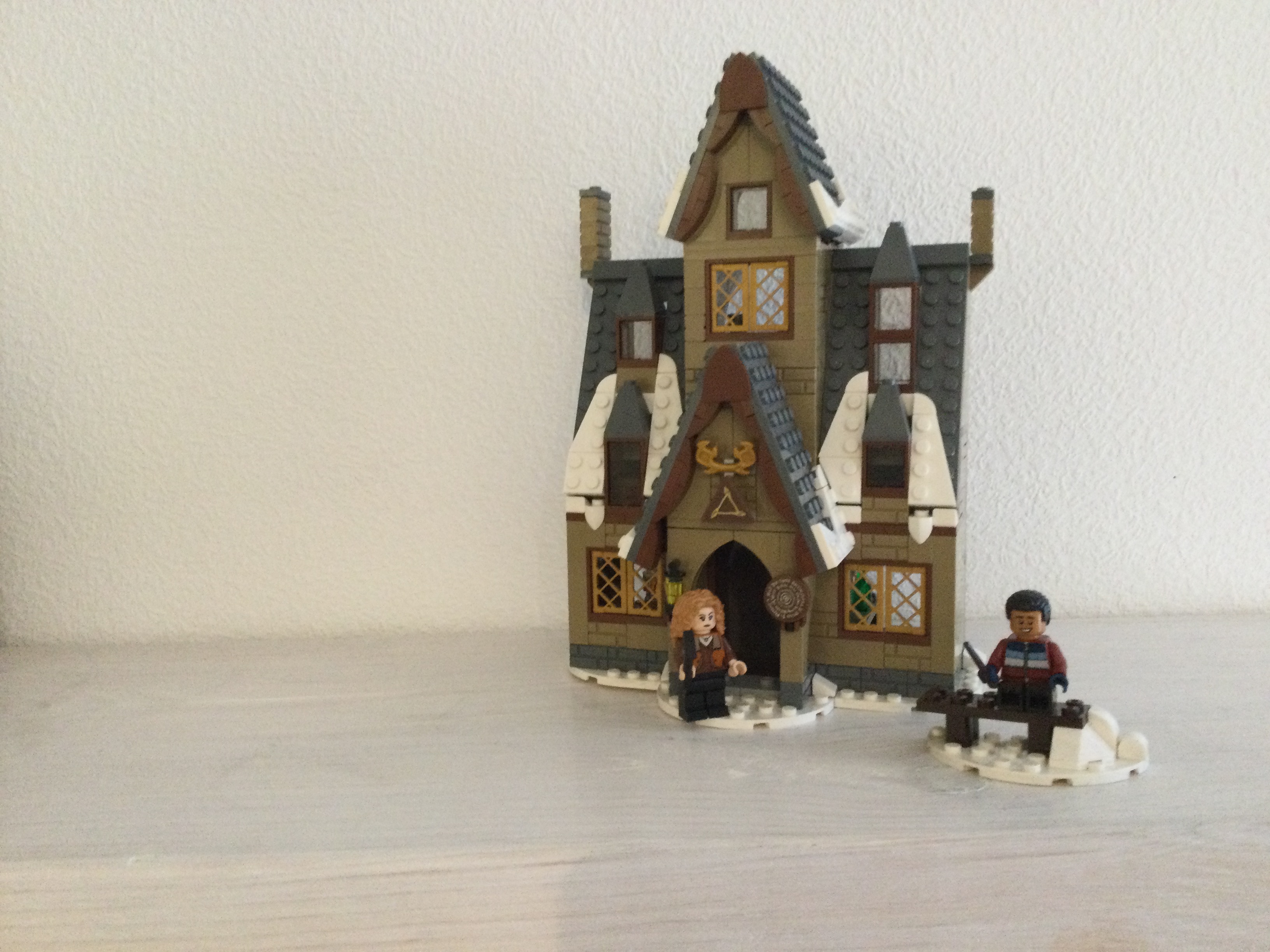}}
    \hfill % \hspace{2cm}
    \subfloat[Stormtrooper]{
        \includegraphics[height=2cm]{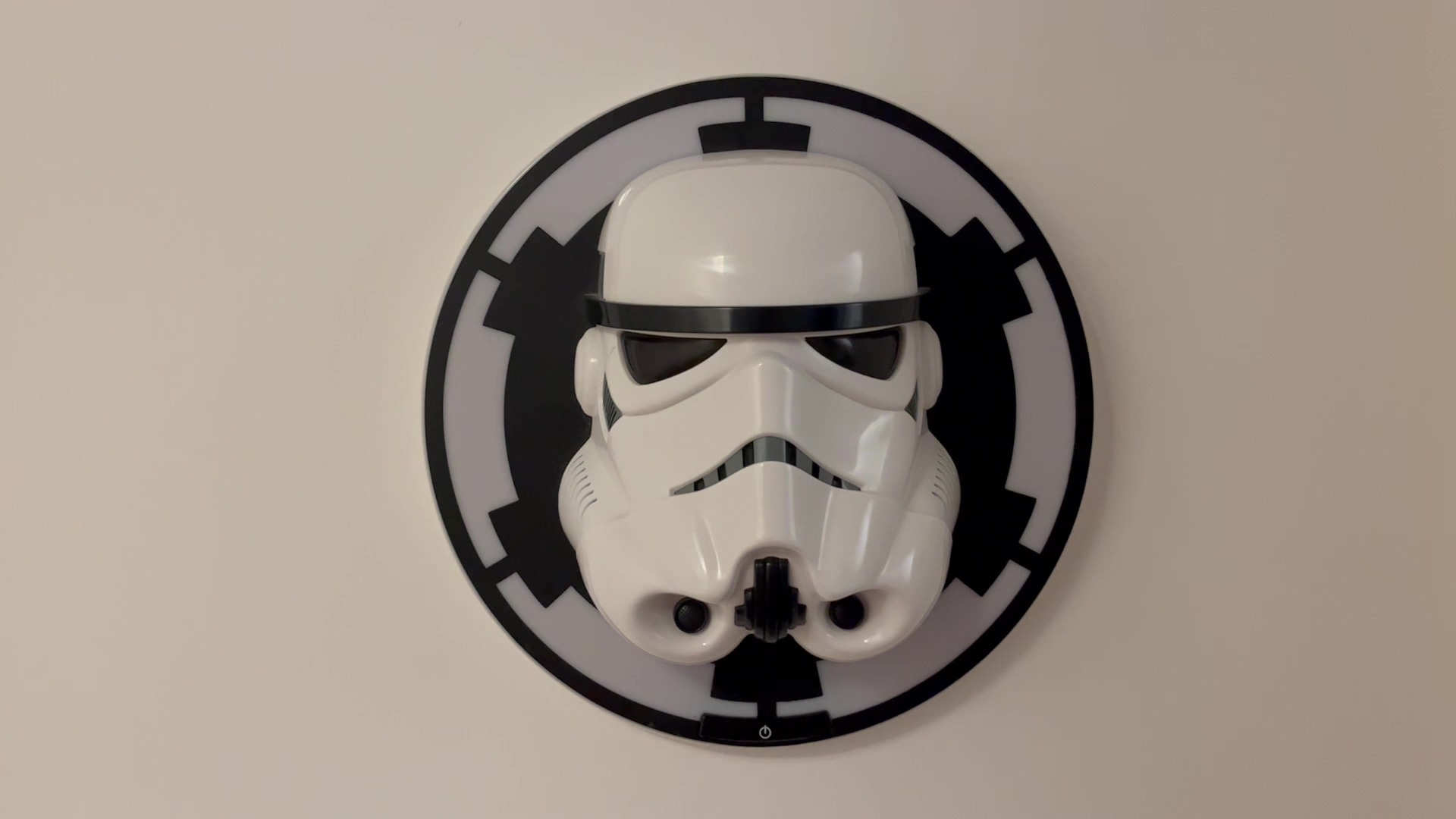}}
    \hfill
    \subfloat[Bike]{
        \includegraphics[height=2cm]{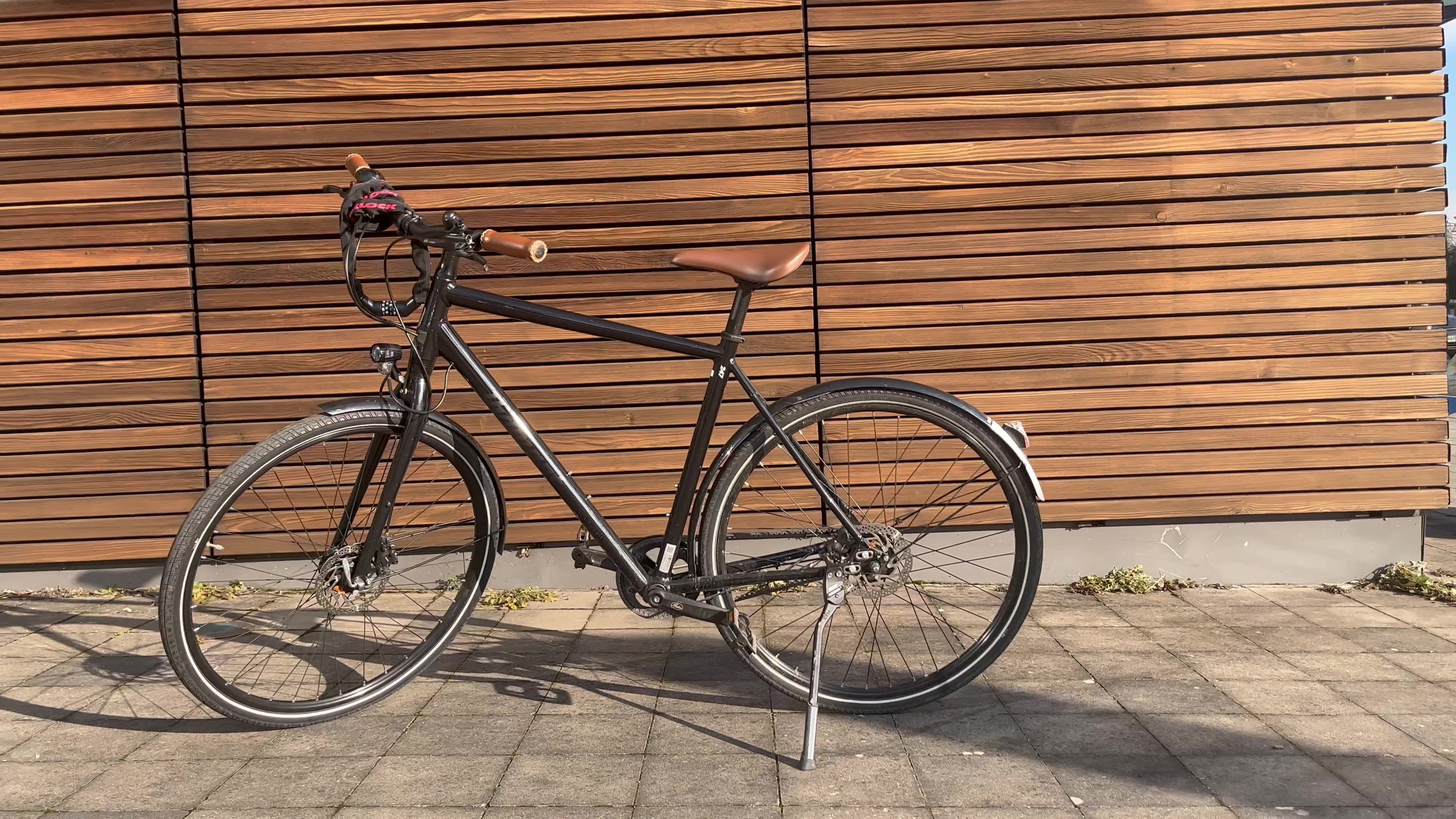}}
    \hfill % \hspace{2cm}
    \subfloat[Fireplug]{
        \includegraphics[height=2cm]{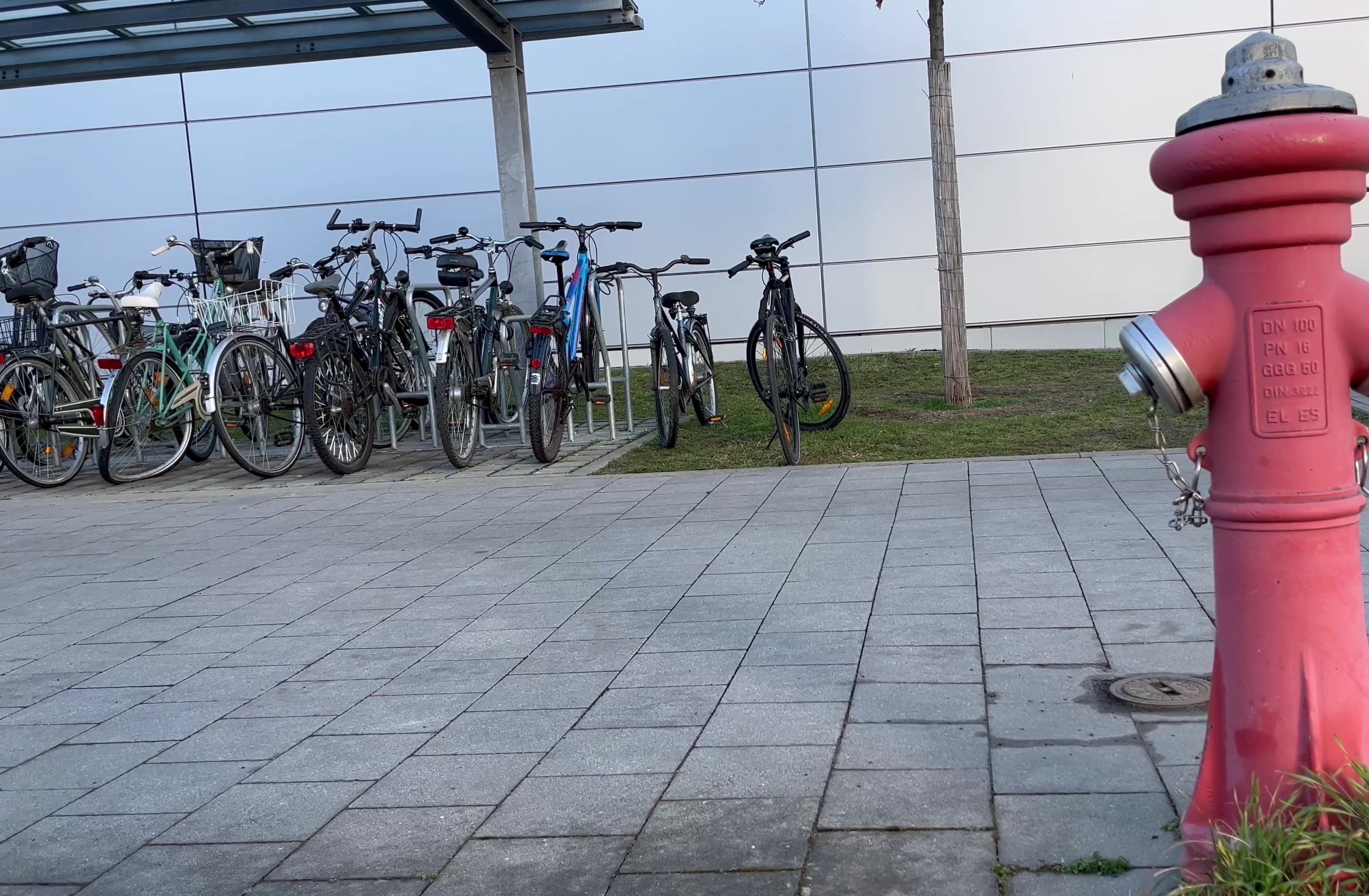}}
    \hfill
    \subfloat[T1]{
        \includegraphics[height=2cm]{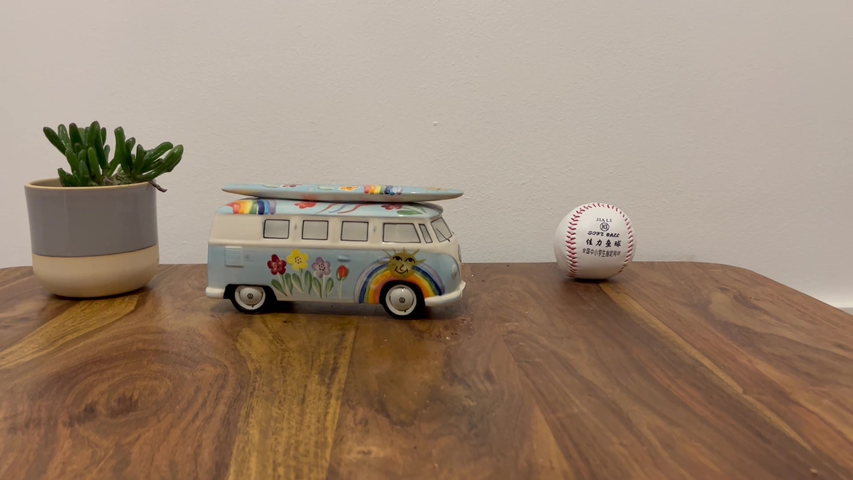}}
        
    \caption{\textbf{Example images from \datasetname.}\label{fig:example_images} We have three indoor and two outdoor scenes. The brick house is captured with an iPad Air 2 and an iPhone 13 mini. The stormtrooper, bike and fireplug are captured with an iPhone 13 mini with different resolutions. The T1 is captured with an iPhone 13 mini and the OAK D-Lite camera.}
\end{figure*}

\section{Dataset}

In our \datasetname dataset we captured \numberofscenes real world scenes. These \numberofscenes scenes are namely T1, brick house, bike, fireplug and stormtrooper. Example images from \datasetname are depicted in Fig.~\ref{fig:example_images}. The scenes are captured with an OAK-D Lite camera with varying resolution, an iPhone mini 13 and an iPad Air 2. We also applied varying resizing factors to later receive more differing intrinsic estimations from COLMAP. The focal lengths of the individual scenes are denoted in Table~\ref{tab:ibleff_license}. All scenes were captured from a video stream. From this video stream we extract one frame per second. This results in the final images of the dataset. To receive the pseudo ground truth we applied COLMAP.

\end{document}